\def\year{2022}\relax
\documentclass[letterpaper]{article} 
\usepackage{aaai22}  
\usepackage{times}  
\usepackage{helvet}  
\usepackage{courier}  
\usepackage[hyphens]{url}  
\usepackage{graphicx} 
\usepackage{natbib}  
\usepackage{caption} 
\DeclareCaptionStyle{ruled}{labelfont=normalfont,labelsep=colon,strut=off} 
\usepackage{algorithm}
\usepackage{algorithmic}

\usepackage{newfloat}
\usepackage{listings}
\floatstyle{ruled}
\newfloat{listing}{tb}{lst}{}
\floatname{listing}{Listing}

\usepackage{cite}
\usepackage{amsmath,amssymb,amsfonts}
\usepackage{algorithmic}
\usepackage{graphicx}
\usepackage{textcomp}
\usepackage{xcolor}
\usepackage{booktabs}
\usepackage{color}
\usepackage[normalem]{ulem}
\usepackage{algorithm,algorithmic}
\usepackage{subcaption}

\usepackage{array}
\usepackage{layouts}
\usepackage{lipsum,graphicx,multicol}
\usepackage{multirow}
\usepackage{pgfplots}
\usepackage{tikz}
\usetikzlibrary{backgrounds}

\pgfplotsset{every axis/.append style={
                    axis x line=middle,    
                    axis y line=middle,    
                    axis line style={<->}, 
                    label style={\small},
                    tick label style={font=\Huge}  
                    }}

\usepackage[labelformat=simple]{subcaption}

\newcolumntype{H}{>{\setbox0=\hbox\bgroup}c<{\egroup}@{}}

\def\blfootnote{\xdef\@thefnmark{}\@footnotetext}

\makeatletter
\def\blfootnote{\xdef\@thefnmark{}\@footnotetext}
\makeatother

\title{Being Friends Instead of Adversaries: \\ Deep Networks Learn from Data Simplified by Other Networks

}
\author {
   Simone Marullo,\textsuperscript{\rm 1,2}
   Matteo Tiezzi,\textsuperscript{\rm 2}
   Marco Gori,\textsuperscript{\rm 2,3}
   Stefano Melacci\textsuperscript{\rm 2}
}
\affiliations {
    \textsuperscript{\rm 1} Dept. of Information Engineering, University of Florence (Italy)\\
    \textsuperscript{\rm 2} Dept. of Information Engineering and Mathematics, University of Siena (Italy)\\
    \textsuperscript{\rm 3} MAASAI, Universit\`{e} C\^{o}te d'Azur,
Nice (France) \\
    simone.marullo@unifi.it, \{mtiezzi,marco,mela\}@diism.unisi.it
}
\begin{document}

\maketitle

\begin{abstract}
Amongst a variety of approaches aimed at making the learning procedure of neural networks more effective, the scientific community developed strategies to order the examples according to their estimated complexity, to distil knowledge from larger networks, or to exploit the principles behind adversarial machine learning.
A different idea has been recently proposed, named Friendly Training, which consists in altering the input data by adding an automatically estimated perturbation, with the goal of facilitating the learning process of a neural classifier. The transformation progressively fades-out as long as training proceeds, until it completely vanishes.
In this work we revisit and extend this idea, introducing a radically different and novel approach inspired by the effectiveness of neural generators in the context of Adversarial Machine Learning. We propose an auxiliary multi-layer network that is responsible of altering the input data to make them easier to be handled by the classifier at the current stage of the training procedure. 
The auxiliary network is trained jointly with the neural classifier, thus intrinsically increasing the ``depth'' of the classifier, and it is expected to spot general regularities in the data alteration process. 
The effect of the auxiliary network is progressively reduced up to the end of training, when it is fully dropped and the classifier is deployed for applications. We refer to this approach as Neural Friendly Training.
An extended experimental procedure involving several datasets and different neural architectures shows that Neural Friendly Training overcomes the originally proposed Friendly Training technique, improving the generalization of the classifier, especially in the case of noisy data. 
\end{abstract}

\section{Introduction}
\label{sec:intro}

\blfootnote{Accepted for publication at the Thirty-Sixth AAAI Conference on Artificial Intelligence (AAAI2022) (DOI: TBA).} 

In the last decade, the scientific research in neural networks studied different aspects of the training procedure, leading to deep neural models of significantly increased quality \cite{batchnormalization,adam,dropout,bengiocurriculum,spcn,zhang2020fat}.
Amongst a large variety of approaches, this paper considers those that are mostly oriented in performing specific actions on the available training data in order to improve the quality of the trained neural classifier.
For example, Curriculum Learning (CL) pursues the idea of presenting the training data in a more efficient manner \cite{bengiocurriculum,Wu2020WhenDC,sinha-curriculum}, exposing the network to simple, easily-discernible examples at first, and to gradually harder examples later, progressively increasing the size of the training set \cite{elman}.
Self-Paced Learning (SPL) \cite{spl-kumar,{spcn}} is another related research area, in which some examples are either excluded from the training set or their impact in the risk function is downplayed if some conditions are met \cite{spcn}.

A common property of CL and SPL is that they essentially sub-select or re-order the training examples, without altering the contents of the data. However, more recently, researches considered approaches that perform transformations of the input data within the input space of the classifier.
Friendly Training (FT) \cite{ft} is a novel approach belonging to the latter category.  
FT allows the training procedure not only to adapt the weights and biases of the classifier, but also to transform the training data in order to facilitate the early fulfilment of the learning criterion. Basically, data are modified to better accommodate the development of the classifier. 
Such transformations (also referred to as ``simplifications'') are controlled and embedded into a precise developmental plan in which the training procedure is progressively constrained to reduce their extent, until data are left in their original version. A key property of FT is that data are altered according to the state of the classifier at the considered stage of the training procedure, and each example is perturbed by a specific offset, obtained by an inner iterative optimization procedure that is started from scratch for each input.
Similarly to CL, the benefits of FT are expected to be mostly evident in the case of noisy examples or in datasets annotated with noisy labels. These are pretty common situations of every data collection process of the real-world. In the case of CL, this has been recently discussed and evaluated in \cite{Wu2020WhenDC}, while in the case of FT the existing evaluation is limited to artificial datasets for digit recognition \cite{ft}. 

\begin{figure}[!ht]
\centering
    \includegraphics[width=0.22\linewidth,angle=0]{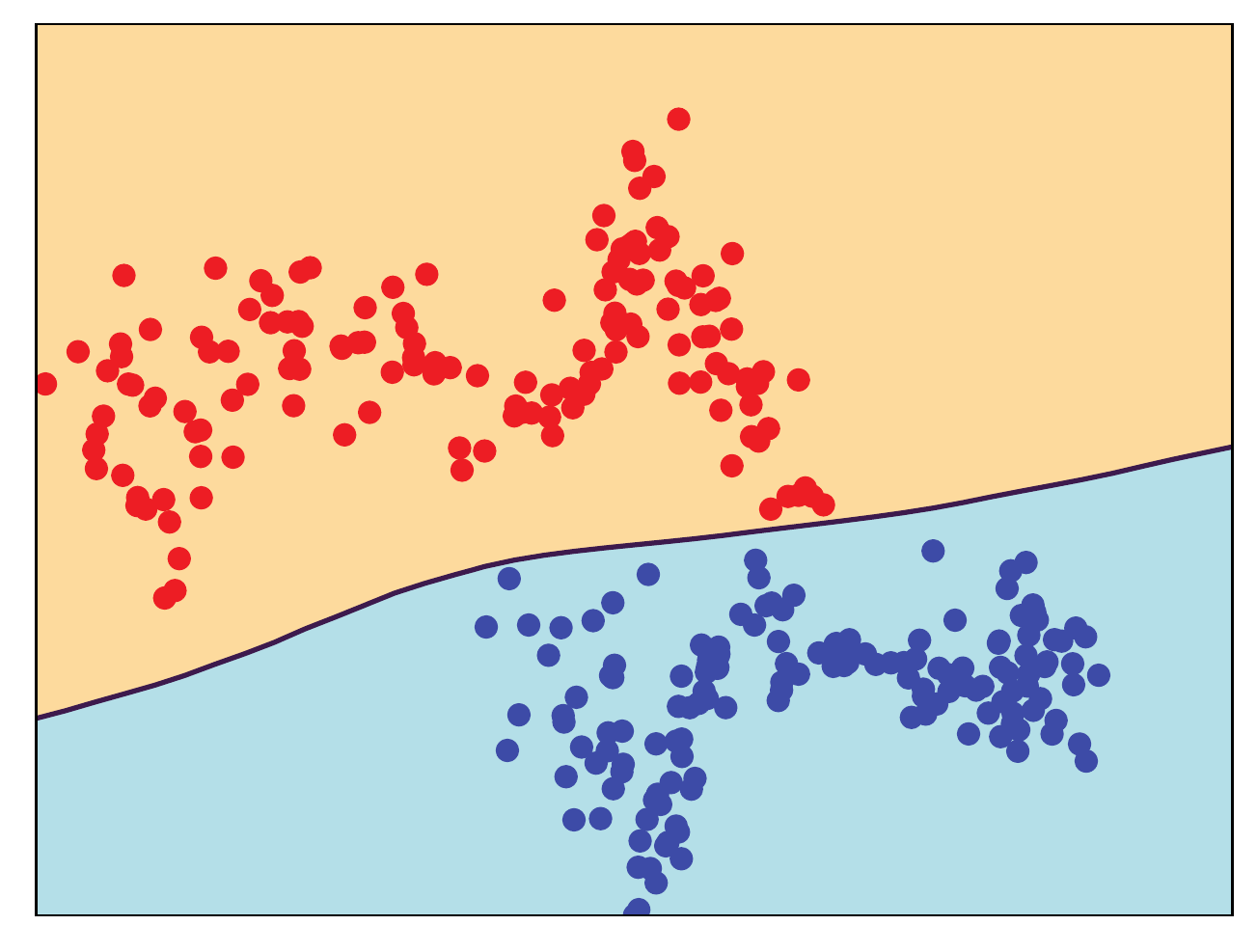}
    \hskip 1mm
    \includegraphics[width=0.22\linewidth,angle=0]{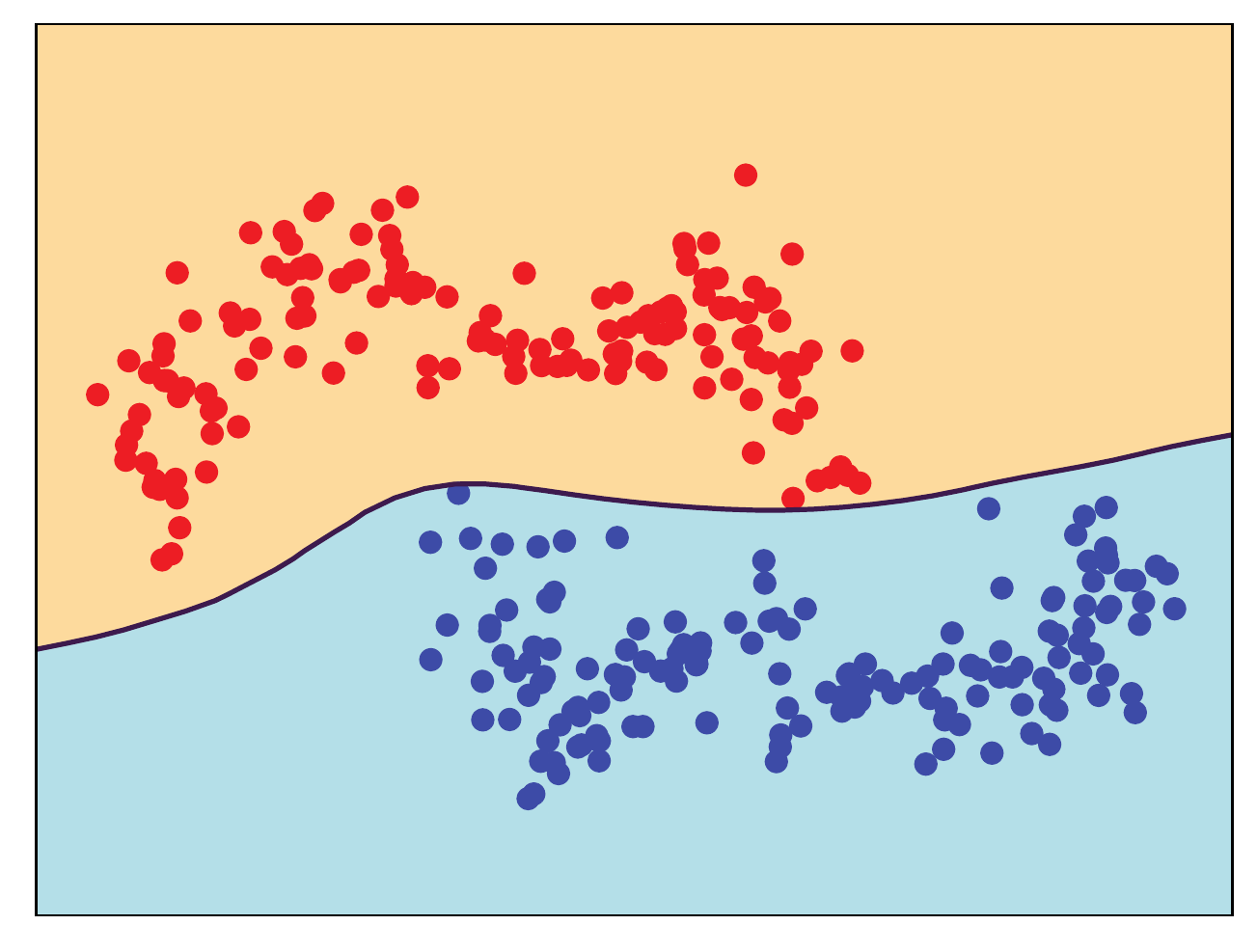}
    \hskip 1mm
    \includegraphics[width=0.22\linewidth,angle=0]{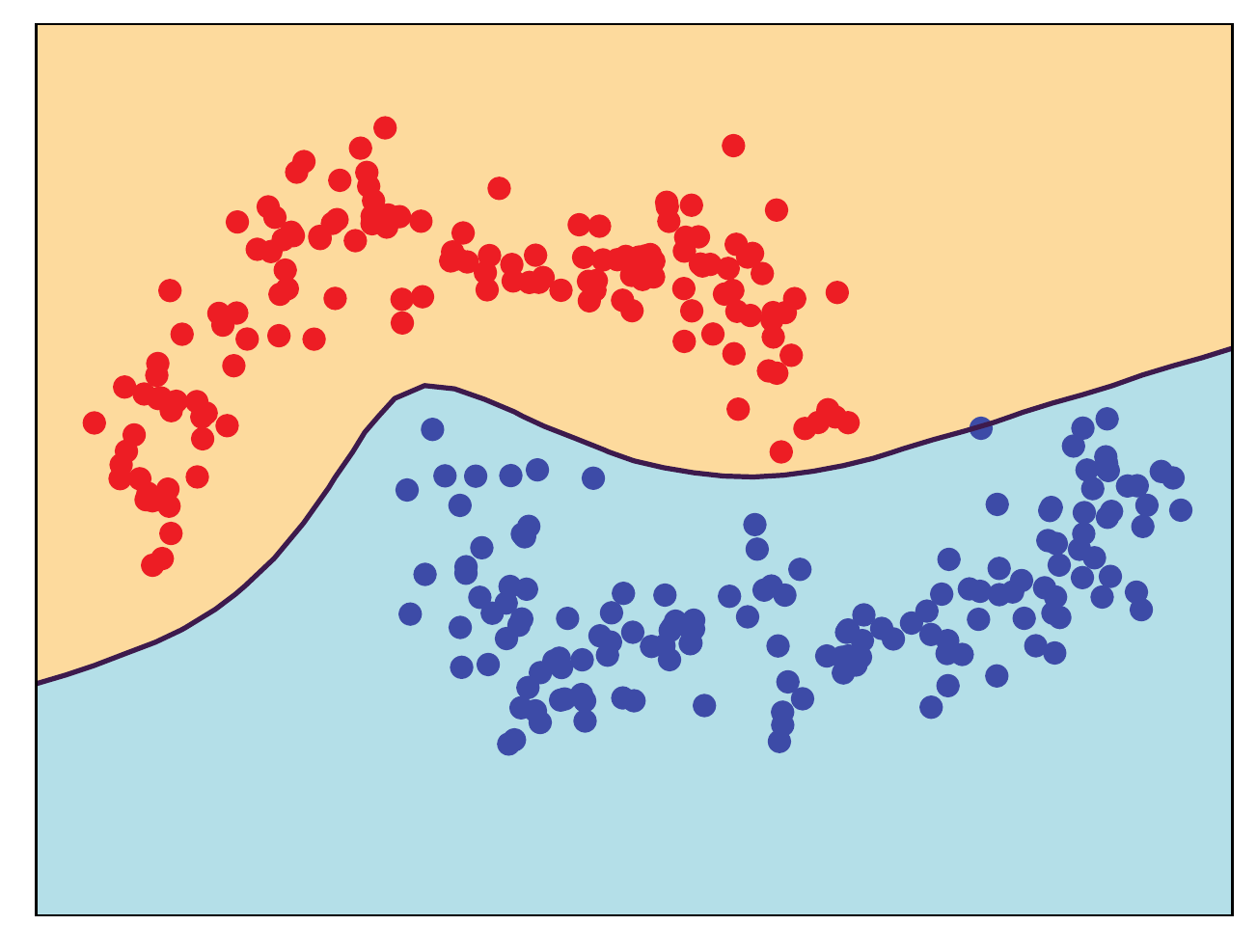}
    \hskip 1mm
    \includegraphics[width=0.22\linewidth,angle=0]{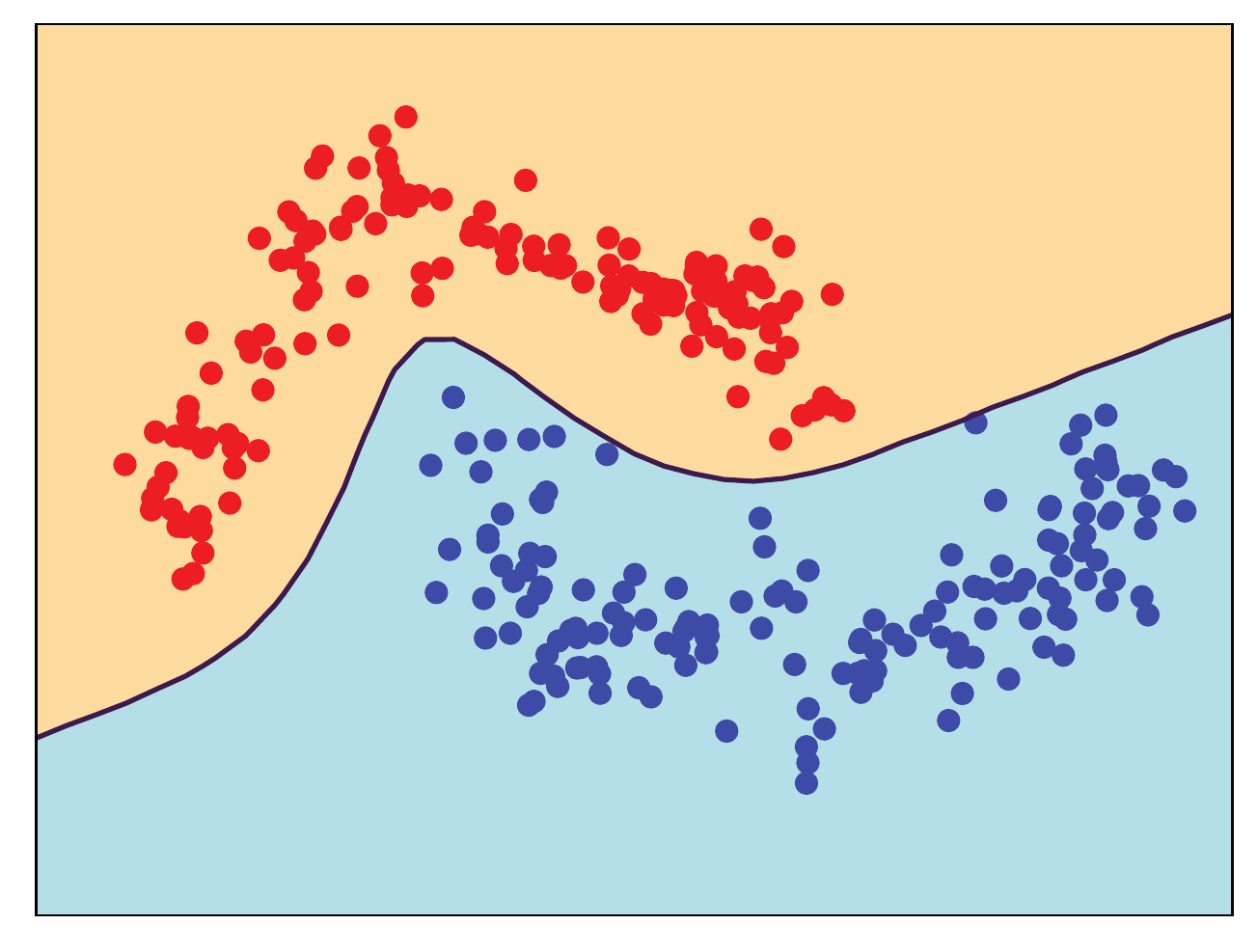}\\
    \vskip 0.5mm
    \includegraphics[width=0.22\linewidth,angle=0]{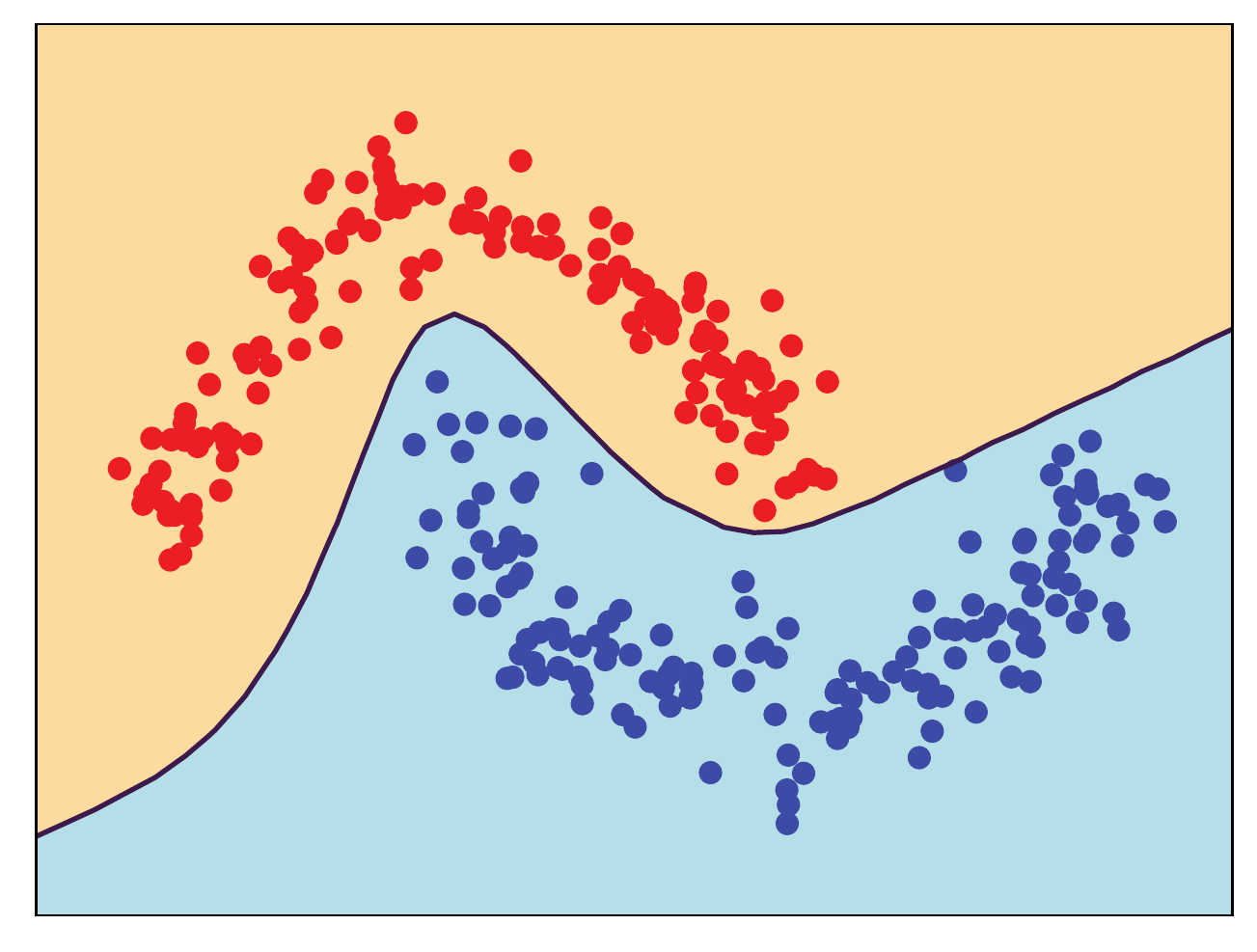}
    \hskip 1mm
    \includegraphics[width=0.22\linewidth,angle=0]{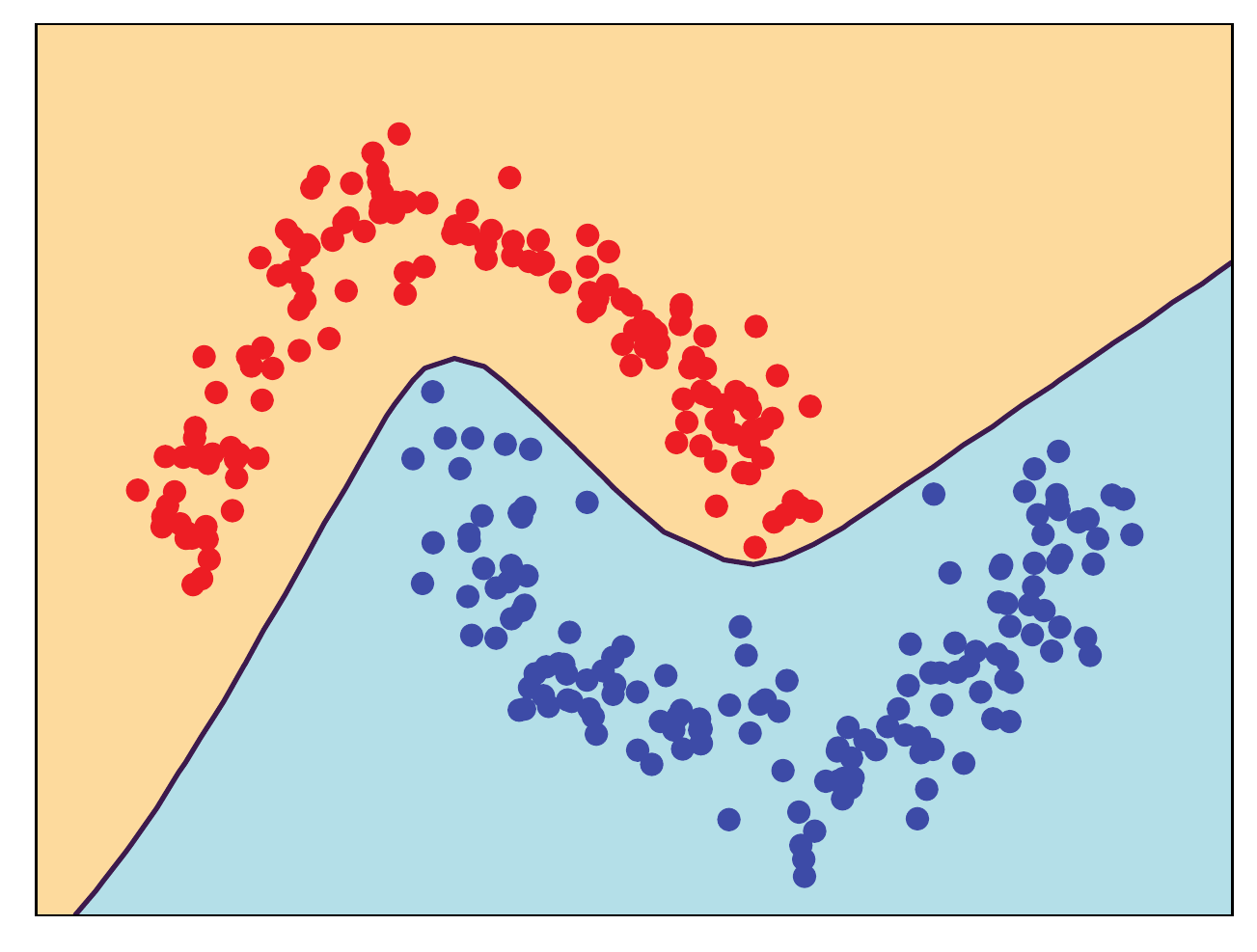}
    \hskip 1mm
    \includegraphics[width=0.22\linewidth,angle=0]{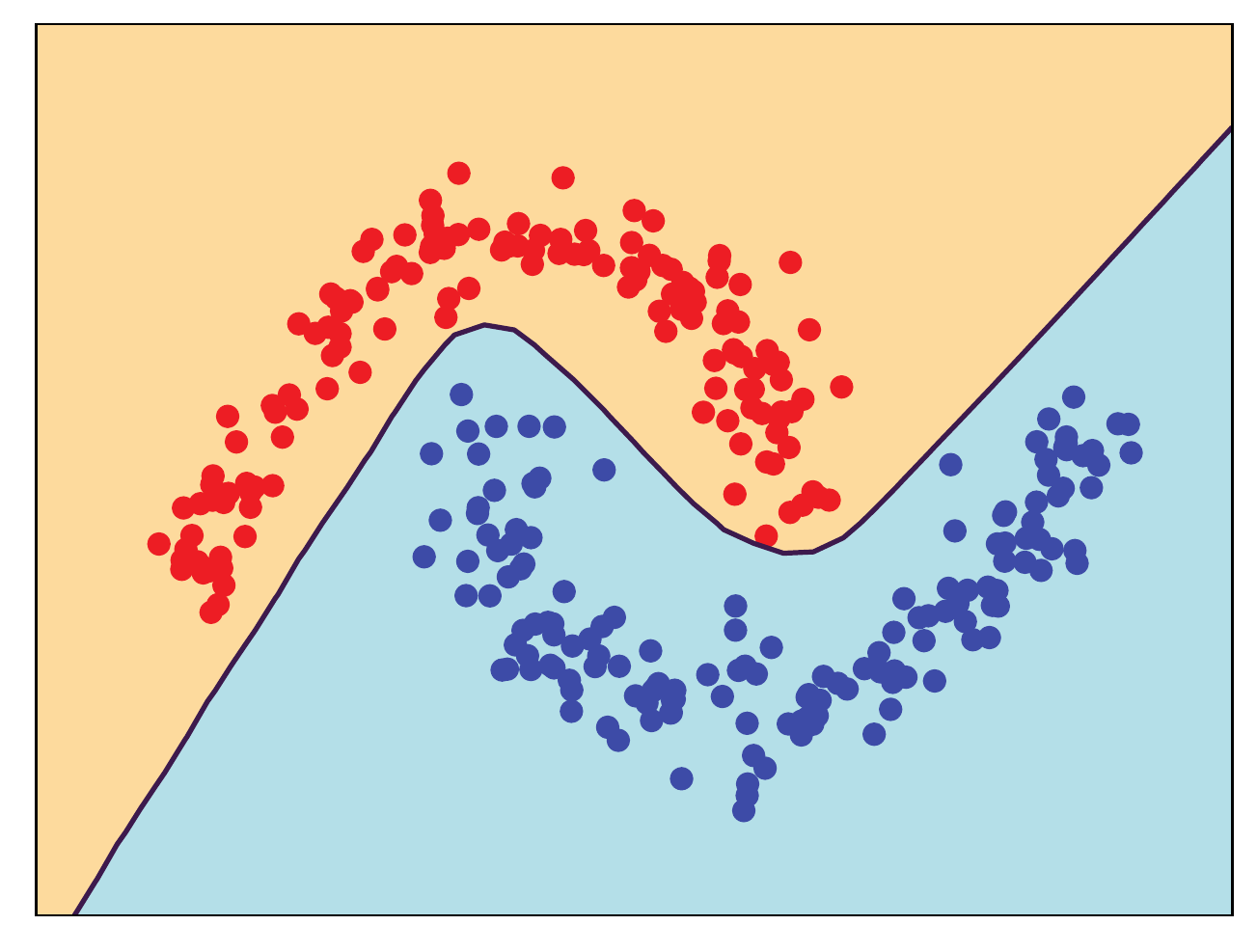}
    \hskip 1mm
    \includegraphics[width=0.22\linewidth,angle=0]{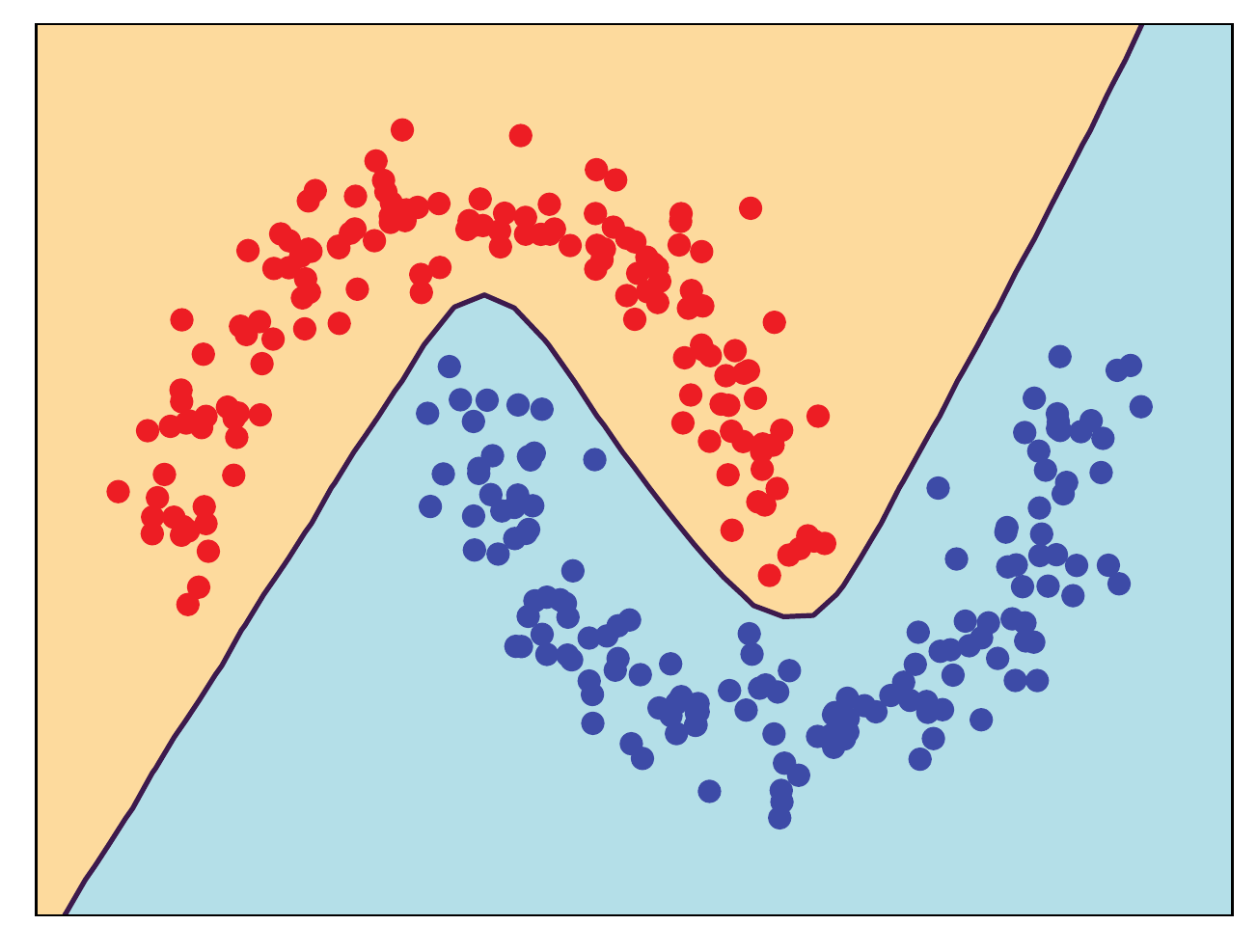}\\
\caption{Left-to-right, top-to-bottom: evolution of the decision boundary developed by a single hidden layer classifier (5 neurons) in the 2-moon dataset, in Neural Friendly Training. Each plot is about a different training iteration ($\gamma$); in the last plot data are not transformed anymore.} 
\label{fig:toy}
\end{figure}

In this paper we revisit and extend the idea of FT, introducing a radically different and novel approach. The intuition behind what we propose is that the data simplification process of FT might include regularities that are shared among different training examples, and that there is an intrinsic coherence in the way data are altered in consecutive training iterations, i.e., similar simplifications might be fine in nearby stages of the training procedure. These considerations are not exploited by FT, which applies an independent perturbation to each example, estimated from scratch at each training step. We propose to introduce an auxiliary multi-layer network, that is responsible of altering data belonging to the input space of the classifier. The auxiliary network is trained jointly with the neural classifier, and it learns how to transform the data to improve the learning process of the classifier itself. The weights of the auxiliary net represent the state of the alteration model, that is progressively updated by the training procedure, thus letting the model evolve as long as time passes.
From an architectural perspective, the auxiliary network extends the classifier by adding a new set of initial layers, thus increasing the ``depth'' of the model.
The effect of the auxiliary network is progressively reduced until the end of training, when it is fully dropped and the classifier is deployed for applications. We refer to this approach as Neural Friendly Training (NFT), and Fig.~\ref{fig:toy} illustrates the behaviour of NFT in a toy 2D classification problem.

Neural models to alter data samples have been proficiently exploited by the Adversarial Machine Learning community \cite{qiu2019semanticadv,ijcai2018-543} with the goal of fooling a classifier. When considering how to improve a classifier exploiting another network, it is immediate to trace a connection also with Knowledge Distillation (KD) \cite{44873,pmlr-v97-phuong19a}, although in KD the main network is supplied with output probability distributions obtained from a pretrained large model. The auxiliary network of NFT learns to transform the input data, closer to what is done by  Spatial Transformer Networks \cite{spatialtransformer} (STN). However, STNs deal with image data only and estimate the parameters of a spatial transformation from a pre-defined family.

The contributions of this paper are: (1) we propose a novel training strategy that allows the machine to simplify the training data by means of an auxiliary network that progressively fades out; (2)  we extend the experimental analysis of the original FT to non-artificial data, and (3) we experimentally compare it with the proposed NFT approach, using 
convolutional and fully connected neural architectures with different numbers of layers. Our results confirm that NFT outperforms FT, proving that NFT is a feasible and effective way to improve the generalization skills of the network and to efficiently deal with noisy training data.

\section{Neural Friendly Training}
\label{sec:method}

We consider a generic classification problem in which we are given a training set $\mathcal{X}$ composed of $n$ supervised pairs, $\mathcal{X} = \{ (x_k, y_k),\ k=1,\ldots,n \}$, being $x_k \in \mathbb{R}^{d}$ a training example labeled with $y_k$.\footnote{We consider the case of classification mostly for the sake of simplicity. The proposed approach actually goes beyond classification problems.}
Given some input data $x$, we denote with $f(x, w)$ the function computed by a neural network-based classifier with all its weights and biases stored into vector $w$. When optimizing the model exploiting a mini-batch based stochastic gradient descent procedure, at each step of the training routine the following empirical risk $L$ measures the mismatch between predictions and the ground truths,
\begin{equation}
    L\left(\mathcal{B}, w \right) = \frac{1}{|\mathcal{B}|} \sum_{i=1}^{|\mathcal{B}|} \ell \left(f\left(x_{i}, w \right) , y_i\right),
    \label{eq:loss}
\end{equation}
where $\mathcal{B} \subset \mathcal{X}$ is a mini-batch of data of size $|\mathcal{B}| \geq 1$, $(x_i, y_i) \in \mathcal{B}$, and $\ell$ is the loss function. Notice that, while we are aggregating the contributes of $\ell$ by averaging over the mini-batch data, every other normalization is fully compatible with what we propose. In the most common case of stochastic gradient optimization, a set of non-overlapping mini-batches is randomly sampled at each training epoch, in order to cover the whole set $\mathcal{X}$. We will refer to what we described so far as Classic Training (CT).  

\paragraph{Friendly Training.} CT provides data to the machine independently on the state of the network and on the information carried by the examples in each $\mathcal{B}$. However, data in $\mathcal{X}$ might include heterogeneous examples with different properties. For instance, their distribution could be multi-modal, it might include outliers or it could span over several disjoint manifolds, and so on and so forth. Existing results in the context of CL \cite{bengiocurriculum,Wu2020WhenDC} and SPL \cite{spcn} (Section~\ref{sec:intro}) show that it might be useful 
to provide the network with examples whose level of complexity progressively increases as long as learning proceeds. However, it is very unlikely to have information on the difficulty of the training examples and, more importantly, if the complexity is determined by humans it might not match the intrinsic difficulty that the machine will face in processing such examples. Alternatively, the value $\ell$ could be used as an indicator to estimate the difficulty of the data, to exclude the examples with largest loss values or to reduce their contribution in Eq.~(\ref{eq:loss}), more closely related to SPL \cite{spl-kumar,spcn}.

Differently from the aforementioned approach, Friendly Training (FT) \cite{ft} \textit{transforms} the training examples according to the state of the learner, with the aim of discarding the parts of information that are too complex to be handled by the network with the current weights, while preserving what sounds more coherent with the expectations of the current classifier.\footnote{This is significantly different from deciding whether or not to keep a training example, to weigh its contribute in Eq.~(\ref{eq:loss}), or to re-order the examples. Interestingly, FT is compatible with (and not necessarily an alternative to) such existing strategies.}
FT consists in alternating two distinct optimization phases, that are iterated multiple times. In the first phase, the training data are transformed in order to make them more easily manageable by the current network. The training procedure must determine how data should be simplified according to the way the current network behaves. In the second phase, the network is updated as in CT, but exploiting the simplified data instead of the original ones. The whole procedure is framed in the context of a developmental plan in which the amount of the alteration is progressively reduced as long as time passes, until it completely vanishes. This is inspired by the basic principle of strongly simplifying the data during the early stages of life of the classifier, in order to favour its development, while the extent of transformation is reduced when the classifier improves its skills. Clearly, to deploy a trained classifier that does not rely on altered data, the impact of the simplification must vanish during the training process, exposing the classifier to the original training data after a certain number of steps.
Formally, FT perturbs the training data by estimating the variation $\delta_i$,
\begin{equation}
    \tilde{x}_i = x_i + \delta_i,
    \label{eq:delta_new}
\end{equation}
for each example $x_i$. Such estimation is repeated from scratch for each training example, and at each training epoch.
    The terms $\delta_i$'s are obtained with the goal of minimizing $L$ in Eq.~(\ref{eq:loss}), replacing $x_i$ with $\tilde{x}_i$ of Eq.~(\ref{eq:delta_new}). Determining an accurate $\delta_i$ might require an iterative optimization procedure, and a maximum number of iterations is defined to control the strength of the perturbation, progressively reduced as long as training proceeds. \footnote{Further details are available in \cite{ft}.} 
    
\begin{figure}[!ht]
    \hskip -1.5mm \includegraphics[width=0.5\textwidth]{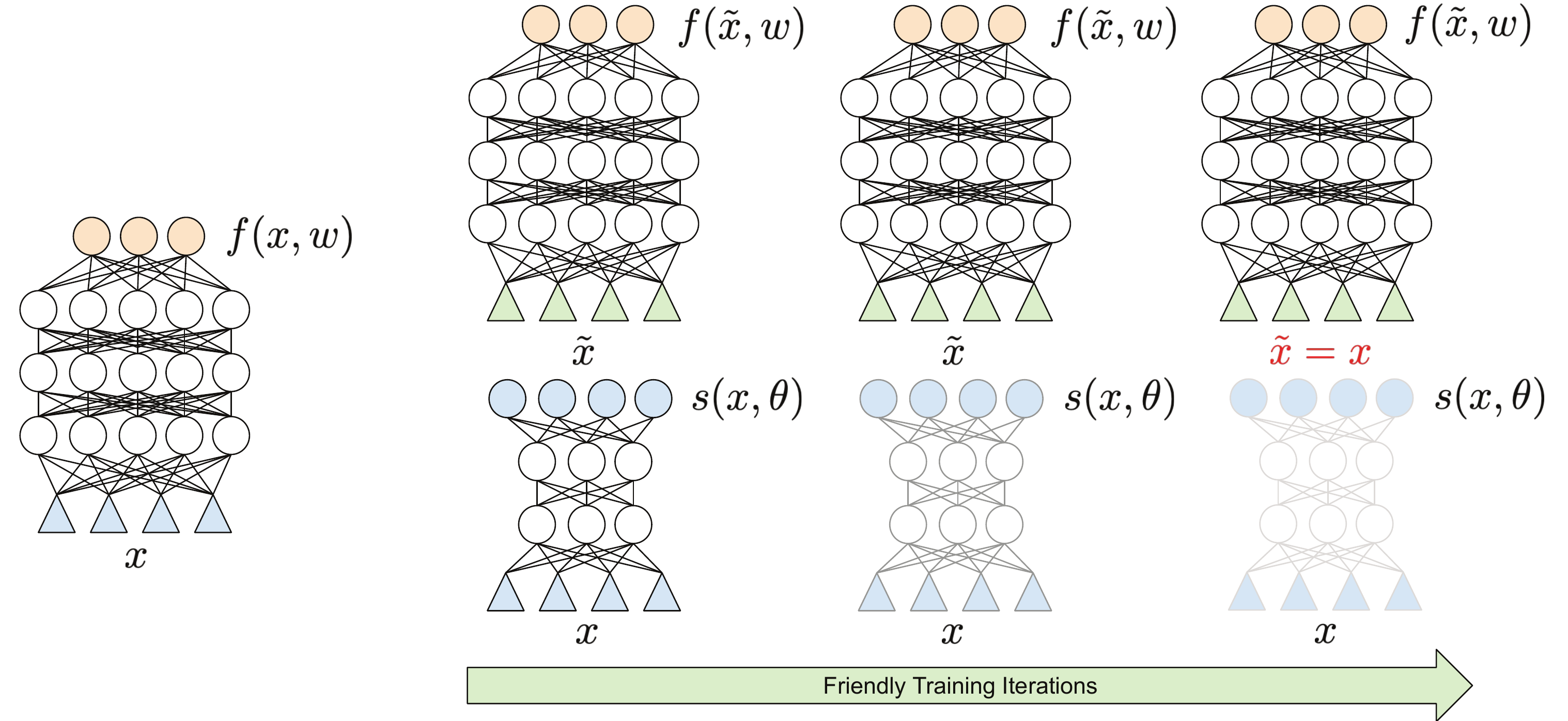}\\
     {\color{white}xxx} (a) \hskip 4cm (b)
    \caption{(a) Classic deep network. (b) Neural Friendly Training (NFT): main deep network (top) and auxiliary network (bottom). The auxiliary network learns how to  simplify the data $x$, while the main network learns the classification task exploiting the simplified data $\tilde{x}$. As long as training proceeds, the effect of the auxiliary network is progressively reduced, until it vanishes (and it is removed).}
    \label{fig:ft}
\end{figure}

\paragraph{Neural Friendly Training.} Despite the novel view introduced by FT, the instance of \cite{ft} is mostly inspired by the basic tools used in the context of Adversarial Training \cite{zhang2020fat}, with a perturbation model that requires a per-example independent optimization procedure. Here we propose to instantiate FT in a different manner, by considering that there might be some regularities in the way data samples are simplified. This leads to the introduction of a more structured transformation function that is shared by all the examples. This intuition is also motivated by recent studies in Adversarial Machine Learning that exploited perturbation models based on generative networks \cite{qiu2019semanticadv,ijcai2018-543}, although with the goal of fooling a classifier. 
Formally, a training sample $x_i\in \mathbb{R}^{d}$ is transformed into $\tilde{x}_i\in \mathbb{R}^{d}$ by means of the function $s(x_i, \theta)$,
\begin{equation}
    \tilde{x}_i = s(x_i, \theta),
    \label{eq:delta}
\end{equation}
being $\theta$ a set of learnable parameters, shared by all the examples.
We consider the case in which $s$ is implemented with an additional neural network, also referred to as \textit{auxiliary network}, whose weights and biases are collected in $\theta$, and we talk about Neural Friendly Training (NFT). For convenience in the notation, we keep the definition of $\delta_i$ inherited from Eq.~(\ref{eq:delta_new}), i.e., $\delta_i = \tilde{x}_i-x_i$.
The term \textit{main network} refers to the network that implements $f$, i.e., the classifier, and we report in Fig.~\ref{fig:ft} a sketch of the proposed model.

In order to setup a valid developmental plan, we introduce an augmented criterion by re-defining the risk $L$ of Eq.~(\ref{eq:loss}),
\begin{eqnarray}
      \hskip -5mm \nonumber   L(\mathcal{B},w,\theta) =   \frac{1}{|\mathcal{B}|} \sum_{i=1}^{|\mathcal{B}|}   \hskip -2mm &\hskip -5mm \Bigg( \hskip -5mm & \hskip -2mm  \ell \big(f ( \underbrace{s({x}_i, \theta)}_{\tilde{x}_i} ,w ) , y_i\big) + \\
      [-3mm] &  & \hskip 5mm \eta \big\| \underbrace{s({x}_i, \theta)  - x_i}_{\delta_i} \big\|^2  \Bigg), 
    \label{eq:loss2}
\end{eqnarray}
where $(x_i,y_i) \in \mathcal{B}$, and $\eta > 0$ is the weight of the squared Euclidean norm of the perturbation $\delta_i$. We indicate with $\gamma \geq 1$ the NFT iteration index, where each iteration consists of the two aforementioned phases. In the first phase, the auxiliary network is updated by minimizing  Eq.~(\ref{eq:loss2}) with respect to $\theta$, keeping the main network fixed.
In the second phase, the auxiliary network has the sole role of transforming the data, while the main network is updated by minimizing  Eq.~(\ref{eq:loss2}) with respect to $w$.  If all the training data is used in this phase, then $\gamma$ boils down to the epoch index (that is the case we considered in the experiments). 
If $\gamma_{max}$ is the maximum number of NFT iterations, we ensure that after $\gamma_{max\_simp} < \gamma_{max}$ steps the data are not perturbed anymore.
In order to progressively reduce the perturbation level, we increase the value of $\eta$ in Eq.~(\ref{eq:loss2}). For a large $\eta$, NFT will strongly penalize the norm of $\delta_i$, becoming the dominant term in the optimization process of the auxiliary network, enforcing the net to keep $\delta_i$ small. 
We indicate with $\eta_{max}$ the maximum possible value of $\eta$, and at each step $\gamma$ of the developmental process we compute $\eta$ using the following law, being $[a]_{+}$ the positive part of $a$,
\begin{minipage}{0.73\columnwidth}
\begin{eqnarray}
\hskip -0.5mm
      \eta = \eta_{max} \hskip -0.5mm \left(\hskip -1mm 1 \hskip -0.5mm - \hskip -0.5mm  \left[ 1 \hskip -0.5mm - \hskip -0.5mm \frac{\gamma \hskip -0.5mm - \hskip -0.5mm 1}{\gamma_{max\_simp} \hskip -0.5mm - \hskip -0.5mm 1} \right]_{+}^2 \hskip -0.5mm \right) \hskip -2mm
    \label{etaplan}
\end{eqnarray} 
\end{minipage}
\hskip 2mm
\begin{minipage}{0.228\columnwidth}
\begin{tikzpicture}[scale=0.27]
\hskip -5mm
\begin{axis}[domain=0:10,samples=1500,
ymin=0, ymax=1.5,
xtick=\empty,ytick=\empty,
extra x ticks=7,extra x tick labels={$\gamma_{max\_simp}$},
extra y ticks=1,extra y tick labels={$\eta_{max}$},
grid=both,axis lines=middle,
 x label style={at={(axis description cs:1.1,0.09), \Huge},anchor=north},
xlabel={$\gamma$},
 y label style={at={(axis description cs:0.0,1.2), \Huge},anchor=north},
ylabel={$\eta$},
]
\addplot+[no marks, line width=1.5pt] {1- (max(1 - (x )/(7) , 0))^2)} node[above left] {};

\end{axis}
\end{tikzpicture}
\end{minipage}
where $\eta \in \left[0, \eta_{max} \right]$.
At $\gamma_{max\_simp}$ iterations, the penalty on $\| \delta_i \|^2$ will reach its maximum weighting. While this enforces the function $s(\cdot, \theta)$ to get closer to the identity function, we have no formal guarantees that it will effectively push the perturbation to zero. For this reason, after $\gamma_{max\_simp}$ iterations we drop the auxiliary network, exposing the system to the original training data. The developmental plan on $\eta$ favours a smooth transition between the setting in which the auxiliary network is used and when it is removed.

The training procedure is detailed in Algorithm~\ref{alg:friendlyn},
\begin{algorithm}
 \caption{Neural Friendly Training.}
 \begin{algorithmic}[1]
 \renewcommand{\algorithmicrequire}{\textbf{Input:}}
 \renewcommand{\algorithmicensure}{\textbf{Output:}}
 \REQUIRE Training set $\mathcal{X}$, initial weights and biases $w$, batch size $b$, max FT steps $\gamma_{max}$, max simplification steps $\gamma_{max\_simp}$, $\eta_{max} > 0$, learning rates $\alpha > 0$ and $\beta > 0$.
 \ENSURE The final $w$.
  \FOR {$\gamma = 1$ to $\gamma_{max}$}
  \STATE Compute $\eta$ following Eq.~(\ref{etaplan})  
  \IF {$\gamma > 1$ \AND $\gamma \leq \gamma_{max\_simp}$}
  \STATE $s$ $\leftarrow$ \texttt{auxiliary\_net}$(\cdot, \theta)$
  \STATE Sample a set of minibatches $B = \{ \mathcal{B}_z \}$ from $\mathcal{X}$
  \FOR {each mini-batch $\mathcal{B}_z \in B$}
  \STATE Compute $\nabla_{\theta} = \frac{\partial L(\mathcal{B}_z,w,h)}{\partial h}\Bigr|_{\substack{h=\theta}}$, see Eq.~(\ref{eq:loss2})
  \hskip 0.0mm \rlap{\smash{$\left.\begin{array}{@{}c@{}}\\{}\\{}\\{}\\{}\end{array}\color{black}\right\}%
  \color{black}\begin{tabular}{l}\hskip -3mm \rotatebox{90}{\scriptsize \textsc{First Phase}: update $s(\cdot, \theta)$}\end{tabular}$}}  
  \STATE $\theta = \theta - \beta \cdot \nabla_{\theta}$
  \ENDFOR
  \ELSE
  \STATE $s \leftarrow I(\cdot)$ 
  \ENDIF
  \STATE Sample a set of minibatches $B = \{ \mathcal{B}_z \}$ from $\mathcal{X}$
\FOR {each mini-batch $\mathcal{B}_z \in B$}
  \STATE Compute $\nabla_{w} = \frac{\partial L(\mathcal{B}_z,h,\theta)}{\partial h}\Bigr|_{\substack{h=w}}$, see Eq.~(\ref{eq:loss2})
     \hskip 1.5mm \rlap{\smash{ $\left.\begin{array}{@{}c@{}}\\{}\\{}\\{}\\{}\end{array}\color{black}\right\}%
  \color{black}\begin{tabular}{l}\hskip -3mm \rotatebox{90}{\scriptsize \textsc{Second Phase}: update $f(\cdot, w)$}\end{tabular}$}} 
  \STATE $w = w - \alpha \cdot \nabla_{w}$
  \ENDFOR  
  \ENDFOR
 \RETURN $w$ 
 \end{algorithmic}
 \label{alg:friendlyn}
 \end{algorithm}
 and in the following lines we provide some further details. The auxiliary network is not updated during the first iteration ($\gamma = 1$), since the main network is still in its initial/random state. After $\gamma_{max\_simp}$ iterations, the auxiliary network is replaced by the identity function $I(\cdot)$ (line 10). Notice that the weight update equations (line 8 and line 16) can include any existing adaptive learning rate estimation procedures, and in our current implementation we are using the Adam optimizer with learning rates $\alpha$ and $\beta$ \cite{adam}, unless differently stated.
 While Algorithm~\ref{alg:friendlyn} formally returns the weights after having completed the last training iteration, as usual, the best configuration of the classifier can be selected by measuring the performance on a validation set (bypassing the auxiliary net at inference time).
 
We qualitatively show the behavior of the proposed training strategy in the toy example of Fig.~\ref{fig:toy}. A very simple network with one hidden layer ($5$ neurons with hyperbolic tangent activation function) is trained on the popular two-moon dataset (two classes, 300 examples), optimized by Adam with mini-batch of size $64$. 
The auxiliary network alters the training data (from the popular 2-moon problem) in order to make them almost linearly separable during the early iterations. Then, the data distribution progressively moves toward the original configuration, and the decision boundary of the main classifier smoothly follows the data. In the last plot, the auxiliary network has been dropped and examples are located at their original positions in final stages of developmental plan.

Of course, NFT increases the complexity of each training step, due to the extra projection computed by the auxiliary network in the forward stage of the classifier and due to the first phase of Algorithm~\ref{alg:friendlyn}. The actual additional computational burden of NFT with respect to CT depends on the architecture of the auxiliary network and on the number of sampled mini-batches. Moreover, instead of Eq.~\ref{etaplan}, different developmental plans could be selected to more quickly reduce the simplification and eventually drop the auxiliary network before the end of training, even if investigating these factors goes beyond the scope of this paper.
When comparing NFT and FT we can see that, from the storage point of view, NFT needs to memorize a new network and the associated intermediate variables for optimization purposes, while FT only requires a new set of variables to store the delta terms. However, from the computational point of view, for each example $x_i$, FT performs $\tau \geq 1$ iterations to update the perturbation $\delta_i$, that implies $\tau$ inference steps on the main network (see Algorithm~1 of \citet{ft}). 
Differently, NFT does not require any inner example-wise iterative procedures (Algorithm~\ref{alg:friendlyn}, first phase). The inference time in the auxiliary network determines the concrete variations in terms of computational times with respect to FT. In our experience, on average, training with NFT took similar times to the ones of FT, since $\tau$ (in FT) gets reduced as time passes and we early stopped the inner FT iterations as suggested in \cite{ft}.

\section{Experiments}
\label{sec:methods}

We carried out a detailed experimental activity aimed at evaluating how NFT behaves when compared to FT. We considered the same experimental conditions of \cite{ft}, initially using the same datasets (Section~\ref{larochelledata}), and then we focused on novel experiences (textual data, Section~\ref{textsux}, pictures of vehicles and animals, Section~\ref{cifar}), where FT was never tested before. We also performed an in-depth analysis on NFT (Section~\ref{sec:in-depth}).

We considered the same four neural classifiers that were used in \cite{ft},\footnote{Code available at \url{https://sailab.diism.unisi.it/friendly}.
}
that consist in two feed-forward Fully-Connected multi-layer perceptrons, referred to as FC-A and FC-B, two Convolutional Neural Networks, named CNN-A and CNN-B, and we also tested a ResNet18 \cite{resnet} in one of the following experiences, motivated by related work \cite{Wu2020WhenDC}.\footnote{FC-A is a simple one-hidden-layer network with hyperbolic tangent activations (10 hidden neurons), while FC-B is deeper and larger model, with 5 hidden layers (2500-2000-1500-1000-500 neurons), batch normalization and ReLU activations. CNN-A consists of 2 convolutional layers, max pooling, dropout and 2 fully connected layers, while CNN-B is deeper (4 convolutional layers). Both of them exploit ReLU activation functions on the convolutional feature maps ($32$-$64$ filters in CNN-A, $32$-$48$-$64$-$64$ filters in CNN-B) and on the fully connected layers ($9216$-$128$ neurons for CNN-A, $5184$-$128$ neurons for CNN-B). Unless differently stated, learning of weights and biases is driven by the minimization of the cross-entropy loss, exploiting the Adam optimizer \cite{adam} with mini-batches of size $32$.}
The auxiliary network was selected depending on the type of data that it is expected to simplify. The output layer has the same size of the input one and linear activation.
In the case of image data (Section~\ref{larochelledata}, \ref{cifar}), the auxiliary network is inspired by U-Net \cite{unet}. U-Net progressively down-samples the image, encoding the context information into the convolutional feature maps, and then it up-samples and transforms the data until it matches the input size, also exploiting skip connections.\footnote{Code: \url{https://github.com/milesial/Pytorch-UNet}.
In the down-sampling part, 2 initial conv. layers encode the image into $n_f$ feature maps. Then, $\nu$ down-sampling blocks (each of them composed of maxpooling and 2 conv. layers) are followed by $\nu$ up-sampling blocks (each of them composed of bilinear upscaling and 2 conv. layers). We considered $\nu \in \{1,2\}$, and $n_f \in \{64,96,128\}$.}
In the case of 1-dim data (Section~\ref{textsux}) we used a fully-connected auxiliary net with $256$ hidden neurons. 

In all the experiments, networks were randomly initialized, providing the exact same initialization to both FT/NFT and CT, and we report results averaged over 3 runs, corresponding to 3 different instances of the initialization process. For each FT/NFT iteration, we sampled non-overlapping mini-batches until all the training data were considered, so that $\gamma$ is also the epoch index. We selected a large number of  epochs $\gamma_{max}$ which we found to be sufficient to obtain a stable configuration of the weights in preliminary experiences (detailed below), and the reported metrics are about the model with the lowest validation error obtained during training. The error rate was selected as the main metric, since it is one of the most common and simple measure in classification problems. 
We performed some preliminary experiments to determine the optimal Adam learning rate in the case of CT. Then, 
we tuned the FT hyper-parameters ($\eta_{max}$,  $\gamma_{max\_simp}$, $\beta$, $n_f$) by grid search (detailed below).
We experimented on two machines equipped with NVIDIA GeForce RTX 3090 (24GB) GPUs.

\subsection{Advanced Digit and Shape Recognition}
\label{larochelledata}

The collection of datasets presented in \cite{mnistvariations} is about $10$-class digit recognition problems and shape-based binary classification tasks ($28\times 28$, grayscale). In detail, \textsc{mnist-rot} consists of MNIST digits rotated by a random angle, while  \textsc{mnist-back-image} features non-uniform backgrounds extracted by some random images, and \textsc{mnist-rot-back-image} combines the factors of variations of the first two datasets.
In \textsc{rectangles-image} we find representations of rectangles, that might be wide or tall, with the inner and outer regions eventually filled with patches taken from other images, while \textsc{convex} is about convex or non-convex white regions on a black background. Datasets ($\approx$ 60k samples) are already divided into training, validation and test set .
We compared the test error rates of the FC-A/B and CNN-A/B models in CT, FT/NFT, and also using the CL-inspired data sorting policy of \cite{ft}, named Easy-Examples First (EEF) that has the same temporal dynamics of FT. 
Experiments are executed for $\gamma_{max} = 200$ epochs, and we selected the model with the lowest validation error considering $\eta_{max} \in \{500 ,1000 ,2000 \}$, $\gamma_{max\_simp} \in \{0.25, 0.5, 0.85\} \cdot \gamma_{max}$, $\beta \in \{ 10^{-5}, 10^{-4}, 5 \cdot 10^{-4}\}$.

Table~\ref{tab:main-table} reports the test error rate of the different models, where other baseline results exploiting different types of classifier can be found in \cite{ft} (typically overcame by FT/NFT). 
Our analysis starts by confirming that the family of Friendly Training algorithms (being them neural or not) very frequently shows better results than CT and of EEF. 
Moreover, the proposed NFT almost always improves the results of FT, supporting the idea of using an auxiliary network to capture regularities in the simplification process. In the case of CNN-A and CNN-B, the error rate of NFT is lower than in FT, with the exception of \textsc{rectangles-image}, where, however, NFT reported a pretty large standard deviation.
In fully-connected architectures FC-A and FC-B, we still observe a positive impact of NFT, that usually beats FT. However, the improvement over CT can be appreciated in a less evident or more sparse manner. As a matter of fact, these architectures are less appropriate than CNNs to handle image data. However, it is still interesting to see how FC-B benefits from the auxiliary network introduced in NFT, that is indeed a convolutional architecture. Overall, results show that using an auxiliary network is better than independently estimating the perturbation offsets of each example, confirming the capability of the network to learn shared facets of the simplification process.

{\setlength{\tabcolsep}{2pt}
\begin{table}[h]
\begin{center}
\scalebox{0.9}{\begin{tabular}{lc|c@{\hspace{1mm}}c@{\hspace{0.5mm}}c@{\hspace{-0.5mm}}c@{\hspace{-0.5mm}}c}
     \toprule
      \multicolumn{2}{c}{$\ $}&mn-back&mn-rot-back&mn-rot&$\ \ $ rectangles $\ \ $&convex\\ \midrule
\multirow{4}{*}{\rotatebox{90}{FC-A}} & CT&$28.34${\tiny $\pm 0.09$}&$64.06${\tiny $\pm 0.31$}&$43.16${\tiny $\pm 0.51$}&$24.31${\tiny $\pm 0.21$}&$33.91${\tiny $\pm 0.44$}\\
& EEF&$\textbf{28.18}${\tiny $\pm 0.47$}&$64.27${\tiny $\pm 0.19$}&$43.91${\tiny $\pm 0.73$}&$24.48${\tiny $\pm 0.11$}&$\textbf{33.17}${\tiny $\pm 0.93$}\\
& FT&$28.66${\tiny $\pm 0.06$}&$64.14${\tiny $\pm 0.36$}&$43.24${\tiny $\pm 0.43$}&$24.64${\tiny $\pm 0.37$}&$34.38${\tiny $\pm 0.22$}\\
& NFT&$\textbf{28.15}${\tiny $\pm 0.04$}&$64.55${\tiny $\pm 0.14$}&$\textbf{42.96}${\tiny $\pm 0.58$}&$24.57${\tiny $\pm 0.19$}&$34.25${\tiny $\pm 1.03$}\\
\midrule 
\multirow{4}{*}{\rotatebox{90}{FC-B}} & CT&$21.06${\tiny $\pm 0.39$}&$51.71${\tiny $\pm 0.79$}&$10.13${\tiny $\pm 0.27$}&$25.10${\tiny $\pm 0.20$}&$27.24${\tiny $\pm 0.05$}\\
& EEF&$21.38${\tiny $\pm 0.18$}&$52.95${\tiny $\pm 0.63$}&$\textbf{10.04}${\tiny $\pm 0.17$}&$\textbf{24.84}${\tiny $\pm 0.32$}&$28.21${\tiny $\pm 0.96$}\\
& FT&$21.74${\tiny $\pm 0.26$}&$\textbf{51.02}${\tiny $\pm 0.07$}&$11.19${\tiny $\pm 0.37$}&$\textbf{24.14}${\tiny $\pm 0.53$}&$27.49${\tiny $\pm 0.07$}\\
& NFT&$\textbf{20.91}${\tiny $\pm 0.52$}&$\textbf{50.20}${\tiny $\pm 0.16$}&$\textbf{10.09}${\tiny $\pm 0.32$}&$\textbf{25.09}${\tiny $\pm 0.09$}&$\textbf{26.81}${\tiny $\pm 0.15$}\\
\midrule
\multirow{4}{*}{\rotatebox{90}{CNN-A}} & CT&$7.25${\tiny $\pm 0.16$}&$29.05${\tiny $\pm 0.45$}&$7.48${\tiny $\pm 0.14$}&$9.86${\tiny $\pm 0.32$}&$8.24${\tiny $\pm 0.09$}\\
& EEF&$\textbf{7.02}${\tiny $\pm 0.08$}&$29.12${\tiny $\pm 0.34$}&$7.61${\tiny $\pm 0.22$}&$12.82${\tiny $\pm 0.70$}&$8.72${\tiny $\pm 0.74$}\\
& FT&$\textbf{6.80}${\tiny $\pm 0.19$}&$\textbf{28.74}${\tiny $\pm 0.29$}&$\textbf{7.36}${\tiny $\pm 0.06$}&$\textbf{9.72}${\tiny $\pm 0.20$}&$8.59${\tiny $\pm 1.44$}\\
& NFT&$\textbf{6.59}${\tiny $\pm 0.09$}&$\textbf{28.67}${\tiny $\pm 0.35$}&$\textbf{7.17}${\tiny $\pm 0.17$}&$10.99${\tiny $\pm 1.89$}&$\textbf{8.03}${\tiny $\pm 0.23$}\\
\midrule
\multirow{4}{*}{\rotatebox{90}{CNN-B}} & CT&$5.15${\tiny $\pm 0.15$}&$23.05${\tiny $\pm 0.21$}&$6.58${\tiny $\pm 0.06$}&$8.10${\tiny $\pm 1.90$}&$3.01${\tiny $\pm 0.41$}\\
& EEF&$\textbf{4.82}${\tiny $\pm 0.19$}&$\textbf{22.89}${\tiny $\pm 0.49$}&$7.02${\tiny $\pm 0.28$}&$8.35${\tiny $\pm 1.01$}&$3.75${\tiny $\pm 0.58$}\\
& FT&$\textbf{5.03}${\tiny $\pm 0.11$}&$\textbf{22.81}${\tiny $\pm 0.36$}&$6.95${\tiny $\pm 0.12$}&$\textbf{7.32}${\tiny $\pm 1.31$}&$\textbf{2.87}${\tiny $\pm 0.42$}\\
& NFT&$\textbf{4.96}${\tiny $\pm 0.34$}&$\textbf{22.22}${\tiny $\pm 0.62$}&$\textbf{6.48}${\tiny $\pm 0.25$}&$\textbf{6.27}${\tiny $\pm 0.62$}&$\textbf{2.78}${\tiny $\pm 0.34$}
\\
\bottomrule
  \end{tabular}}
\end{center}
\caption{Comparison of different classifiers (FC-A, FC-B, CNN-A, CNN-B) and learning algorithms (CT, EEF, FT from \cite{ft} and our NFT) -- datasets of Section~\ref{larochelledata} (where \textsc{mn} stands for \textsc{mnist} and removing the suffix \textsc{image}). Test error and standard deviation over 3 runs are reported. For each architecture, those results that improve the CT case are in bold. 
}
\label{tab:main-table}
\end{table}}

\subsection{Sentiment Analysis}
\label{textsux}

We investigate how NFT behaves in Natural Language Processing considering the task of Sentiment Analysis (positive/negative polarity). We selected two datasets and considered different representations of the examples. The first dataset is \textsc{imdb} \cite{maas-EtAl:2011:ACL-HLT2011}, also known as Large Movie Review Dataset, that is a collection of $50$k highly-polar reviews from the IMDB database. We considered a vocabulary of the most frequent $20$k words and TF-IDF \cite{tfidf-jones} representation of each review.
The second dataset, \textsc{wines} \cite{thoutt_wine_nodate}, collects  $130$k wine reviews scored in $[80,100]$, that we divided into two classes, i.e., $[80,90)$ vs. $[90,100]$. In this case, 
in order to acquire a broader outlook on the effect of NFT, we chose a different text representation, exploiting a pretrained Transformer-based architecture (DistilRoBERTa \cite{reimers-2019-sentence-bert}, with average pooling to compute dense representations of size $768$ for each review.
We trained the deeper fully-connected architecture, FC-B, for $30$ epochs. 
NFT hyper-parameters were selected in $\eta_{max} \in \{10,100,500,1000,2000\}$, $\gamma_{max\_simp} \in \{0.05, 0.1, 0.25, 0.85, 0.5\} \cdot \gamma_{max}$, $\beta \in \{ 10^{-5}, 10^{-4}, 5 \cdot 10^{-4} \}$. Concerning FT, we extended the grids of \cite{ft} for all the new experiences of this paper, testing further parameter configurations (supplementary material at \url{https://sailab.diism.unisi.it/friendly}).

As reported in Table~\ref{tab:otherdata} (top), the performance of CT is consistently improved by NFT, achieving lower error rates in  both the datasets (and representations).
The sentence classification task appears to be slightly more difficult in \textsc{wines}. This is probably due to the fact that wine reviews are less polarized, being them all highly scored. 
Concerning \textsc{imdb}, the superiority of NFT over CT and also FT is evident. Overall, these results confirm the versatility of NFT.
\begin{table}[!hb]
\scalebox{0.9}{\begin{tabular}{l|c}
\toprule
\multicolumn{1}{c}{$\ $} &imdb\\ \midrule
FC-B CT&$13.27$ {\tiny $\pm 0.19$}\\
FC-B FT&${13.66}$ {\tiny $\pm 0.69$}\\
FC-B NFT \hskip 1.7mm $\ $ &$\textbf{11.93}$ {\tiny $\pm 0.09$}\\
\bottomrule
\end{tabular}}
\hskip 1mm
\scalebox{0.9}{\begin{tabular}{l|c}
\toprule
\multicolumn{1}{c}{$\ $} &$\ \ \ \ \ $wines$\ \ \ \ \ $\\ \midrule
FC-B CT&$17.38$ {\tiny $\pm 0.15$}\\
FC-B FT&$\textbf{17.07}$ {\tiny $\pm 0.11$}\\
FC-B NFT \hskip 0.8mm $\ $ &$\textbf{17.15}$ {\tiny $\pm 0.12$}\\
\bottomrule
\end{tabular}}\\
\vskip 1mm
\scalebox{0.9}{\begin{tabular}{l|c}
\toprule
\multicolumn{1}{c}{$\ $} &cifar-10\\ \midrule
CNN-B CT&$29.75$ {\tiny $\pm 0.37$}\\
CNN-B FT&$30.19$ {\tiny $\pm 0.53$}\\
CNN-B NFT&$\textbf{29.00}$ {\tiny $\pm 0.36$}
\\
\bottomrule
\end{tabular}}
\hskip 1mm
\scalebox{0.9}{\begin{tabular}{l|c}
\toprule
\multicolumn{1}{c}{$\ $} &cifar-10-n10\\ \midrule
ResNet CT&$9.30$ {\tiny $\pm 0.16$}\\
ResNet FT&$\textbf{8.92}$ {\tiny $\pm 0.23$}\\
ResNet NFT&$\textbf{8.10}$ {\tiny $\pm 0.19$}
\\
\bottomrule
\end{tabular}}
\caption{Comparison of classifiers with different architectures and learning algorithms (CT, FT) -- data of Section~\ref{textsux} (top) and Section~\ref{cifar} (bottom). Mean test error is reported with standard deviation. Results improving CT are in bold.}\label{tab:otherdata}
\end{table}

\subsection{Image Classification}
\label{cifar}
CIFAR-10 \cite{krizhevsky_learning_2009} is a popular Image Classification dataset, consisting of $60$k $32 \times 32$ color images from 10 different classes.
We divided the original training data into training and validation sets ($10$k examples used as validation set), and we initially evaluated NFT using the previously described generic CNN-B architecture. Table \ref{tab:otherdata} (bottom-left) shows that while were not able to improve the results CT using FT, NFT slightly improves the quality of the network, reducing the error rate and further confirming its benefits. 

\begin{figure}[!hb]
\includegraphics[width=\columnwidth]{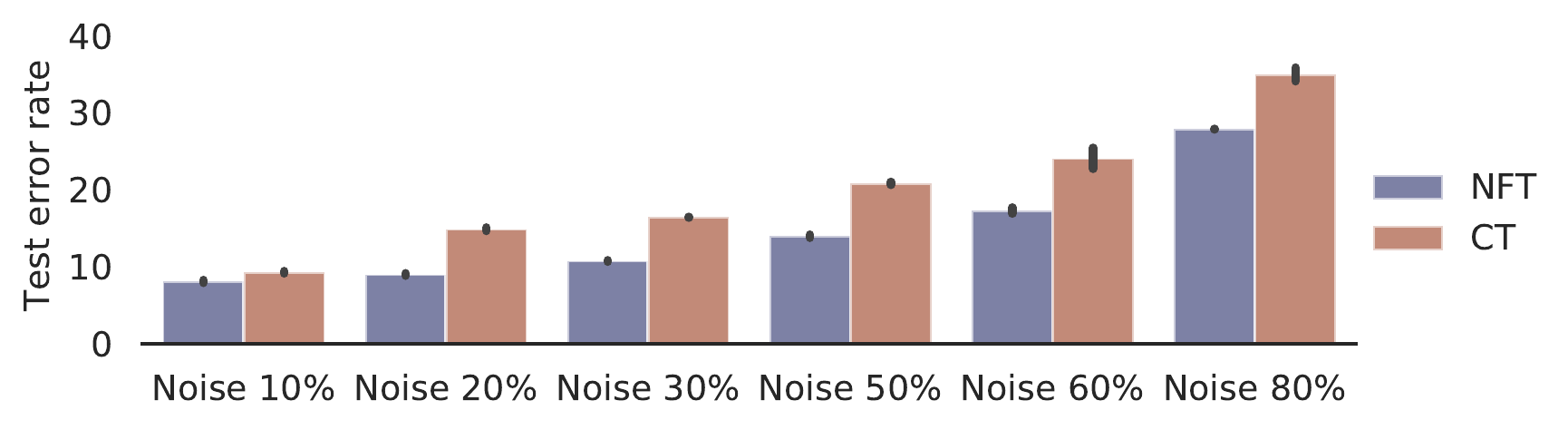}
\centering
\caption{ResNet18 on \textsc{CIFAR-10} dataset for different amounts of noisy labels. Error bars include standard dev.}
\label{fig:histo-noisy}
\end{figure}

However, state-of-the art convolutional networks specifically designed/tuned for CIFAR-10 usually achieve lower error rates, so that we decided to perform a more specific experimental activity.
In particular, we considered ResNet18 \cite{resnet}, inheriting all the carefully selected optimization parameters and tricks that yield state-of-the-art results in CIFAR-10.\footnote{Stochastic Gradient Descent (learning rate $0.1$ with cosine annealing learning rate scheduler) with momentum ($0.9$) and weight decay ($5 \cdot 10^{-4}$), mini-batches of size 128, data augmentation -- see \url{https://github.com/kuangliu/pytorch-cifar}.}
Since FT/NFT bring marginal benefits over CT, we designed a more challenging condition following the setup of recently published CL activity \cite{Wu2020WhenDC}. We introduced some noise by randomly permuting 10\% of the target labels, generating what we will refer to as \textsc{cifar-10-n10}.
We trained the network for $250$ epochs, and reported results in Table~\ref{tab:otherdata} (bottom-right). NFT hyper-parameters were selected in $\eta_{max} \in \{500,1000,2000\}$, $\gamma_{max\_simp} \in \{0.25, 0.5, 0.7\} \cdot \gamma_{max}$, $\beta \in \{ 10^{-4}, 5 \cdot 10^{-4} \}$. The learning rate scheduler is applied starting from $\gamma_{max\_simp}$ with an initial learning rate which is $0.1 \cdot \alpha$.
We observe that NFT effectively helps also when dealing with this type of network. While FT also carries a small improvement, it is far from the one obtained by NFT. We further investigated this result by varying the amount of noise injected into the training labels. Fig.~\ref{fig:histo-noisy} compares CT and NFT for different noise levels, up to $80\%$. Interestingly, the impact of NFT becomes more and more evident, gaining $\approx 8\%$ in strongly noisy environments, confirming that data simplification helps the main network to better discard the noisy information. 

\subsection{In-Depth Analysis}\label{sec:in-depth}
\begin{figure}
\centering
\begin{minipage}{0.48\textwidth}
    \hskip 8mm $x\ \ \ \ \ \ \ \ \ $ $\delta\ \ \ \ \ \ \ \ \ $ $\tilde{x}$ \hskip 11mm $x\ \ \ \ \ \ \ \ \ \ $ $\delta\ \ \ \ \ \ \ $ $\tilde{x}$\\
    \rotatebox{90}{\hskip 10mm FT}
    \centering
    \includegraphics[width=0.38\textwidth]{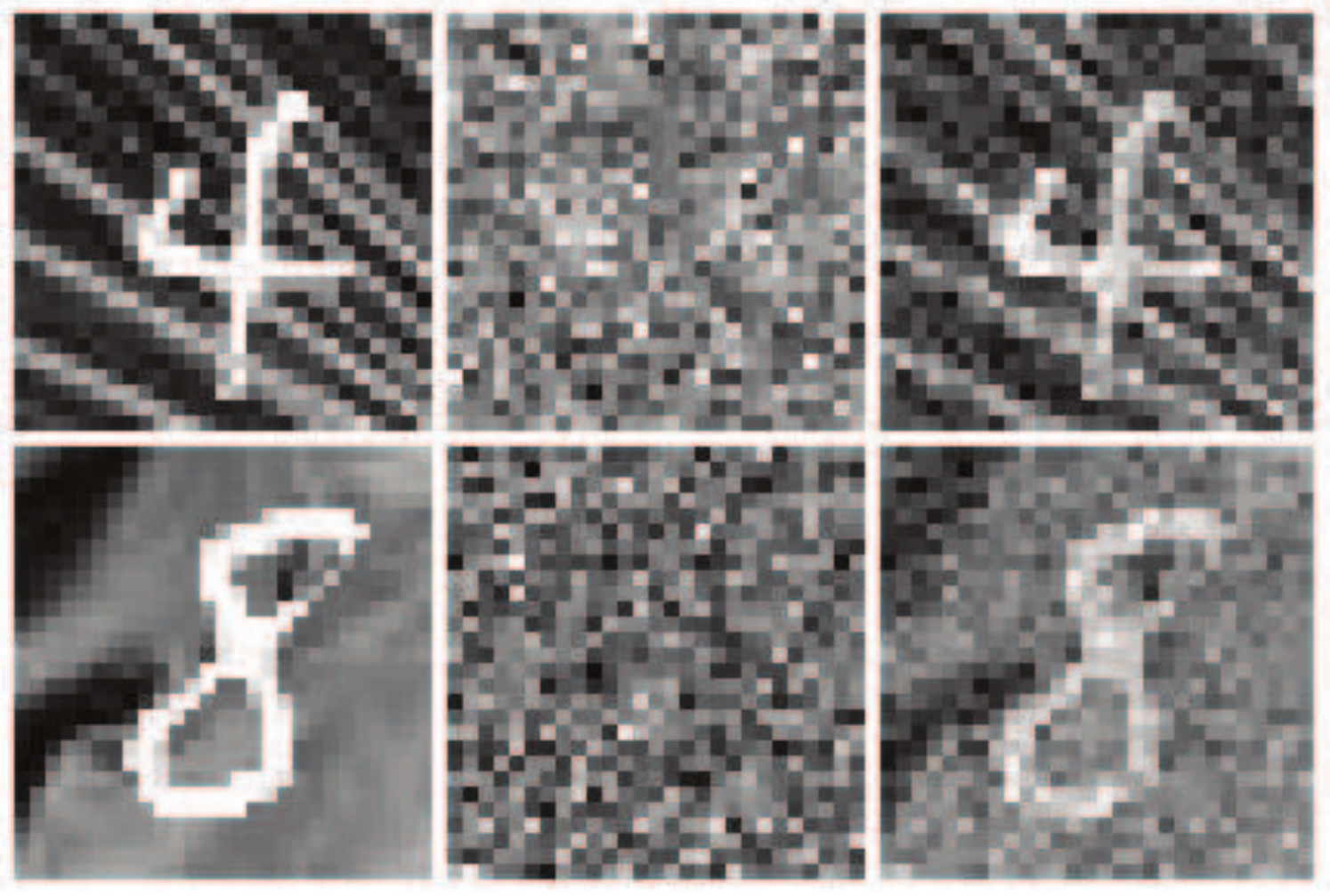}
    \hskip 1mm
    \includegraphics[width=0.38\textwidth]{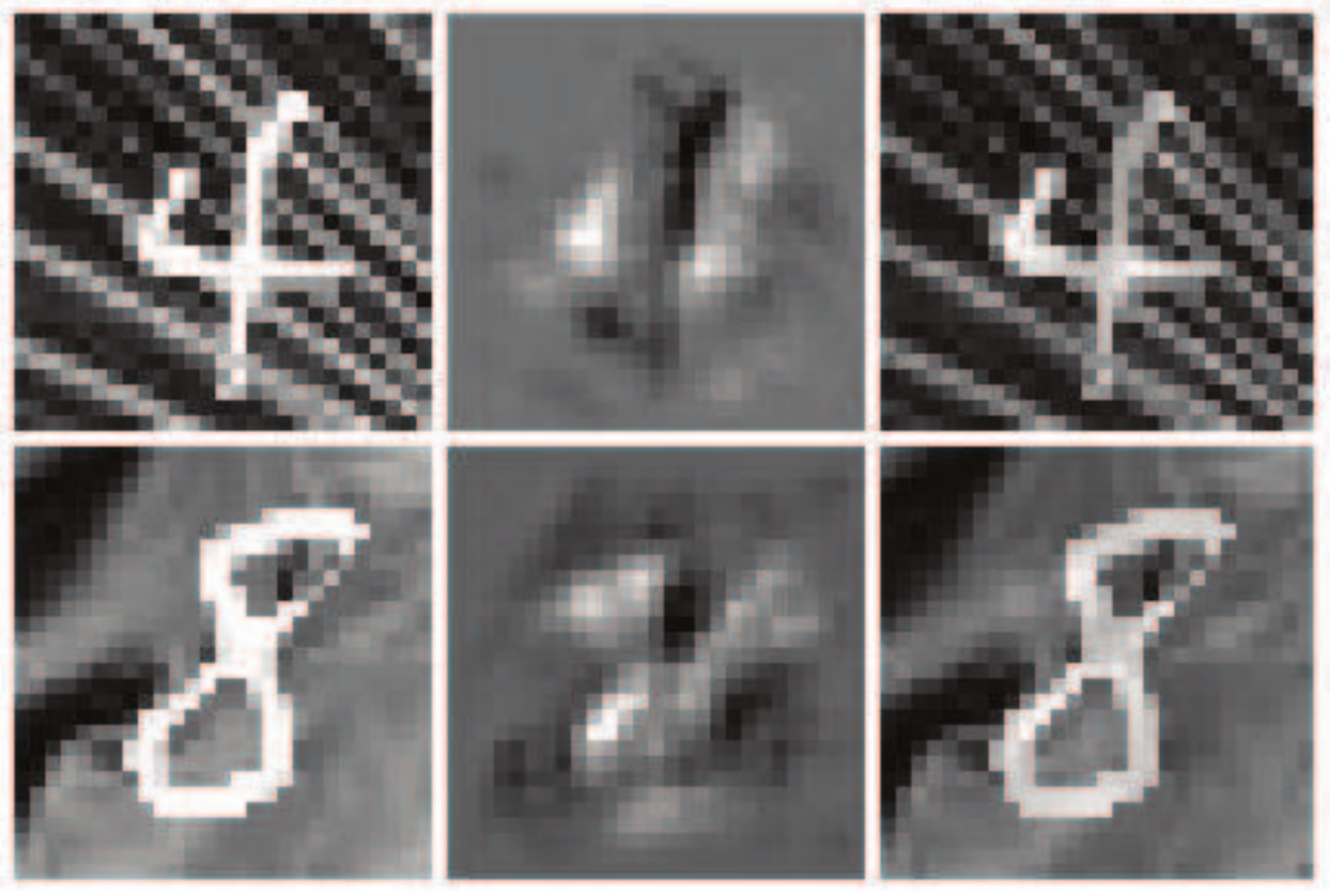}\\
    \hskip 2mm $x\ \ \ \ \ \ \ \ \ $ $\delta\ \ \ \ \ \ \ \ \ $ $\tilde{x}$ \hskip 11mm $x\ \ \ \ \ \ \ \ \ \ $ $\delta\ \ \ \ \ \ \ $ $\tilde{x}$\\
    \rotatebox{90}{\hskip 8.5mm NFT}
    \includegraphics[width=0.38\textwidth]{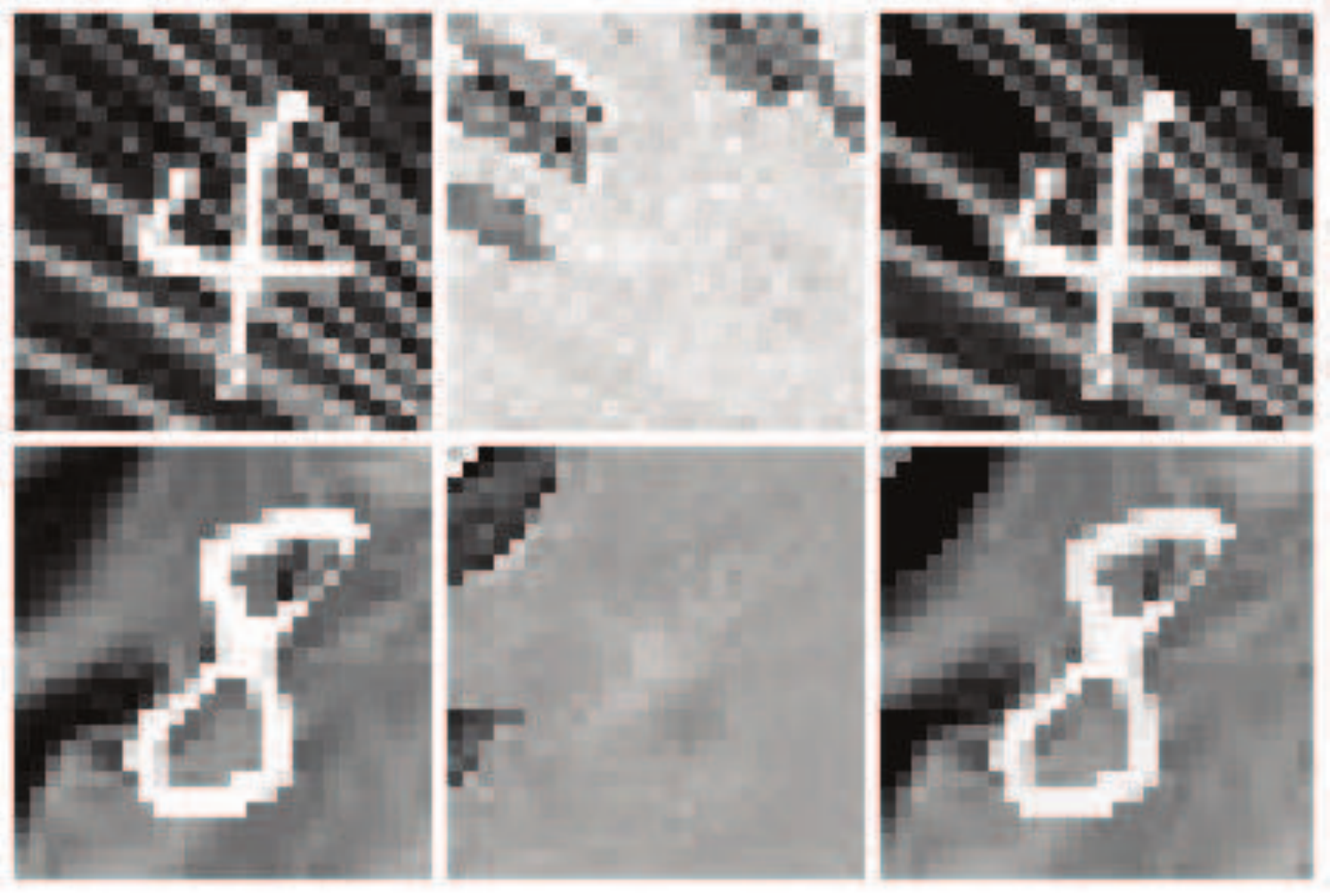}
    \hskip 1mm
    \includegraphics[width=0.38\linewidth]{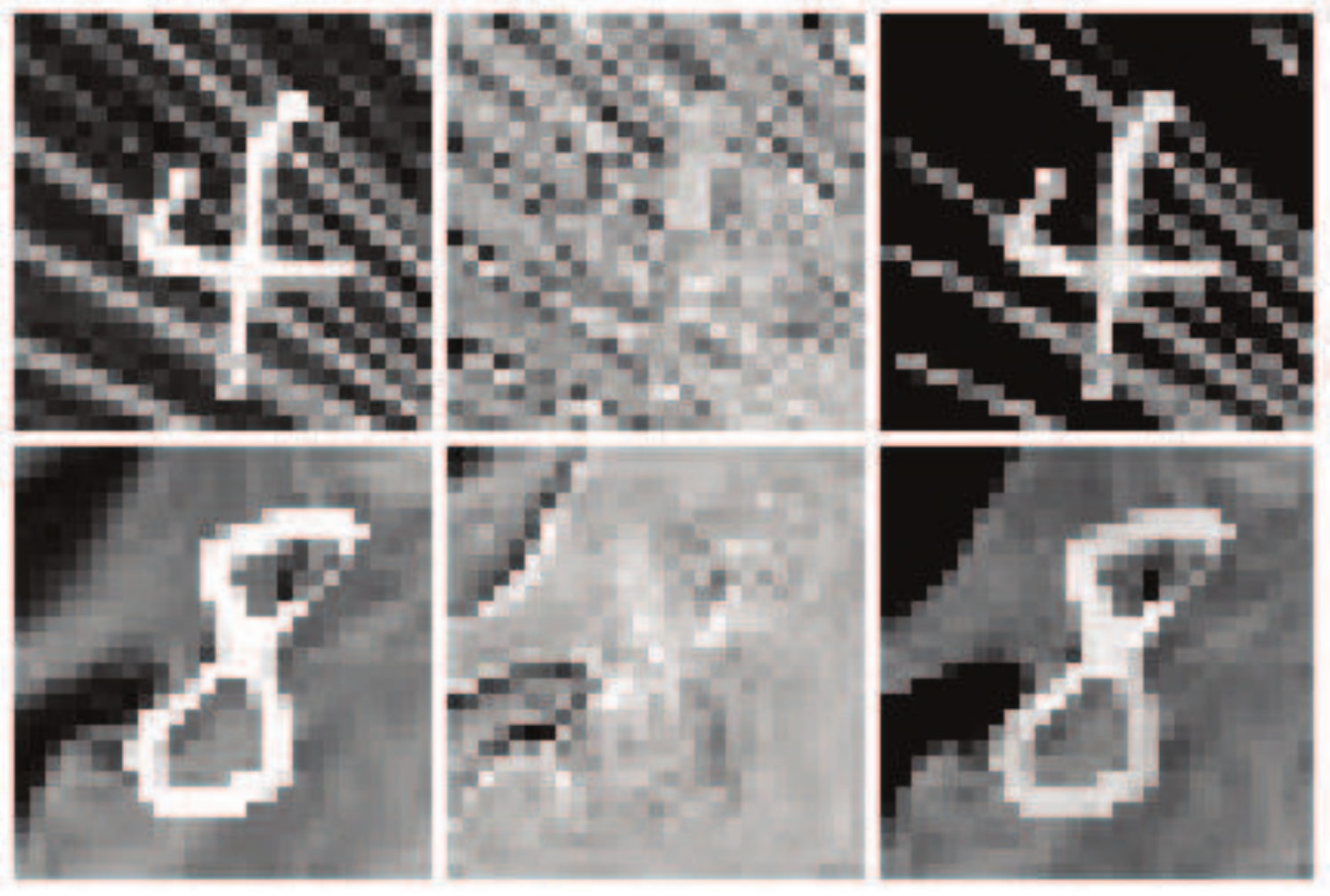}\\
    \hskip 4.5mm FC-A\hskip 2.3cm CNN-A
\end{minipage}
\caption{\textsc{mnist-back-image}. Original data $x$ , perturbation $\delta$ (normalized) and resulting ``simplified'' images $\tilde{x}$ for FC-A and CNN-A at the end of the $1$st epoch. Some simplifications are hardly distinguishable. Top: FT. Bottom: NFT.}
\label{fig:simpl-images}
\end{figure}

We qualitatively compared NFT and FT in the \textsc{mnist-back-image} dataset of Section~\ref{larochelledata}, in which the important information is known (the digits), since the background is uncorrelated with the target. We mostly considered the CNN-A model, for which NFT led to the most significant improvements with respect to CT (Table~\ref{tab:main-table}).
In Fig.~\ref{fig:simpl-images} we show how examples are affected when using an auxiliary network (bottom - NFT) or when independent transformations are estimated for each example through a gradient-based procedure (top - FT).
Estimating the transformation function with a neural model leads to qualitatively different behavior. 
We observe that FT yields structured perturbations only when paired with CNN-A, emphasizing the digit areas. Differently NFT shows more natural  perturbation patterns, removing distracting cues (background).
Basically, the convolutional auxiliary net leads to transformations with much more detailed awareness of the visual structures.

\begin{figure}[!ht]
\centering
\includegraphics[width=0.29\textwidth]{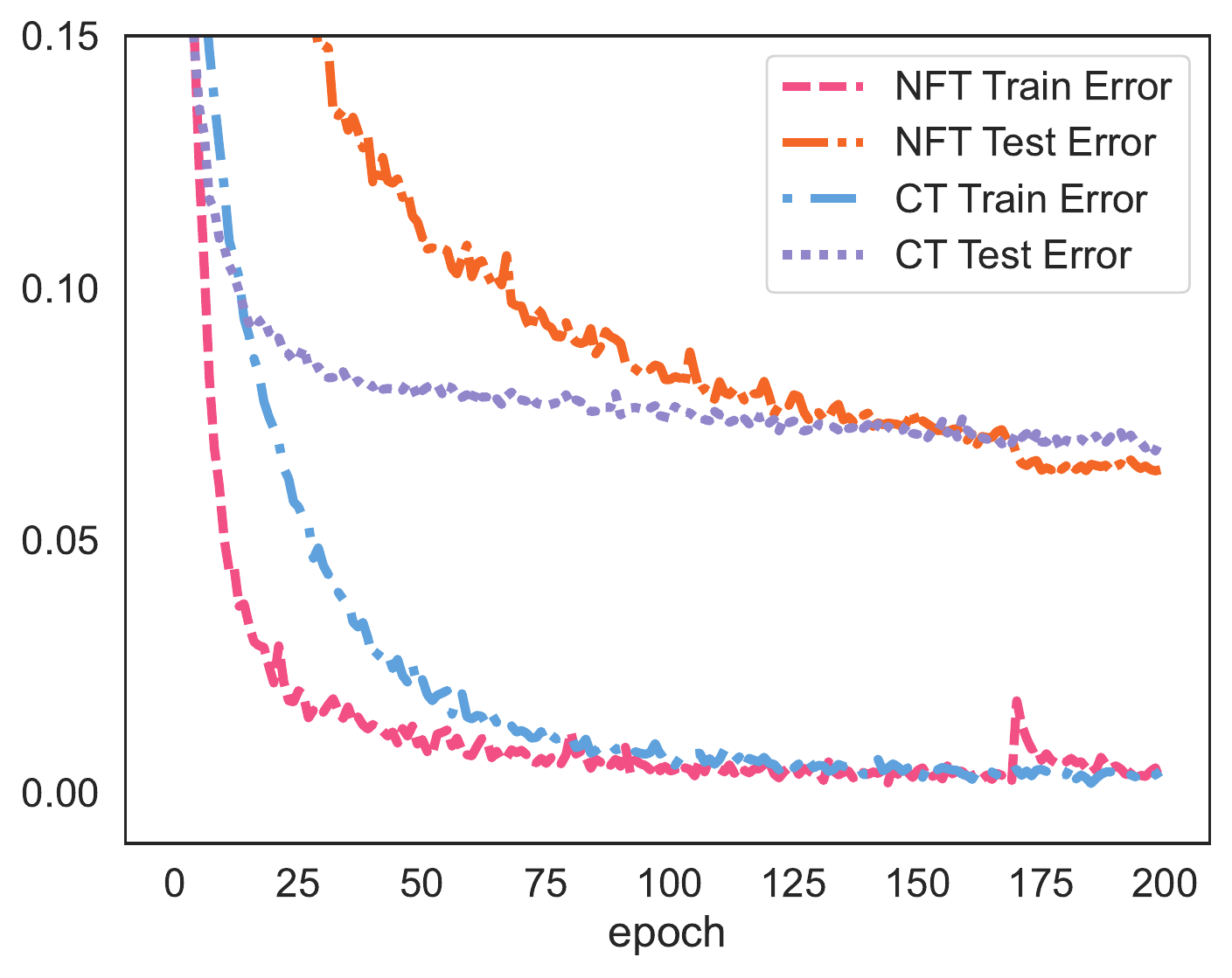}
\caption{Training and test error rates for NFT and CT on a single run -- \textsc{mnist-back-image}, CNN-A (best viewed in colors). The auxiliary network is dropped at epoch $175$. The training error of NFT is initially lower than in the case of CT since the auxiliary network simplifies the data. Differently, the test error is initially larger, since the test set is not simplified. As training proceeds, the simplification vanishes and the test data become aligned with the training ones.}
\label{histoleft}
\end{figure}

In Fig.~\ref{histoleft}, we report the evolution of test error rate during the training epochs (\textsc{mnist-back-image}, CNN-A), comparing NFT and CT. The developmental plan reduces the impact of the perturbation until epoch $175$ (afterwards, data are not altered anymore).
The small bump right before such epoch is due to the final transition from altered to original data. The test error of NFT is higher than the one of CT when data are altered, as expected, while it becomes lower when the auxiliary network is dropped. On the other hand, fitting training data is easier during the early epochs in NFT, due to the simplification process.

\begin{figure}[!ht]
\begin{center}
\begin{minipage}{0.45\textwidth}
\includegraphics[width=0.48\textwidth]{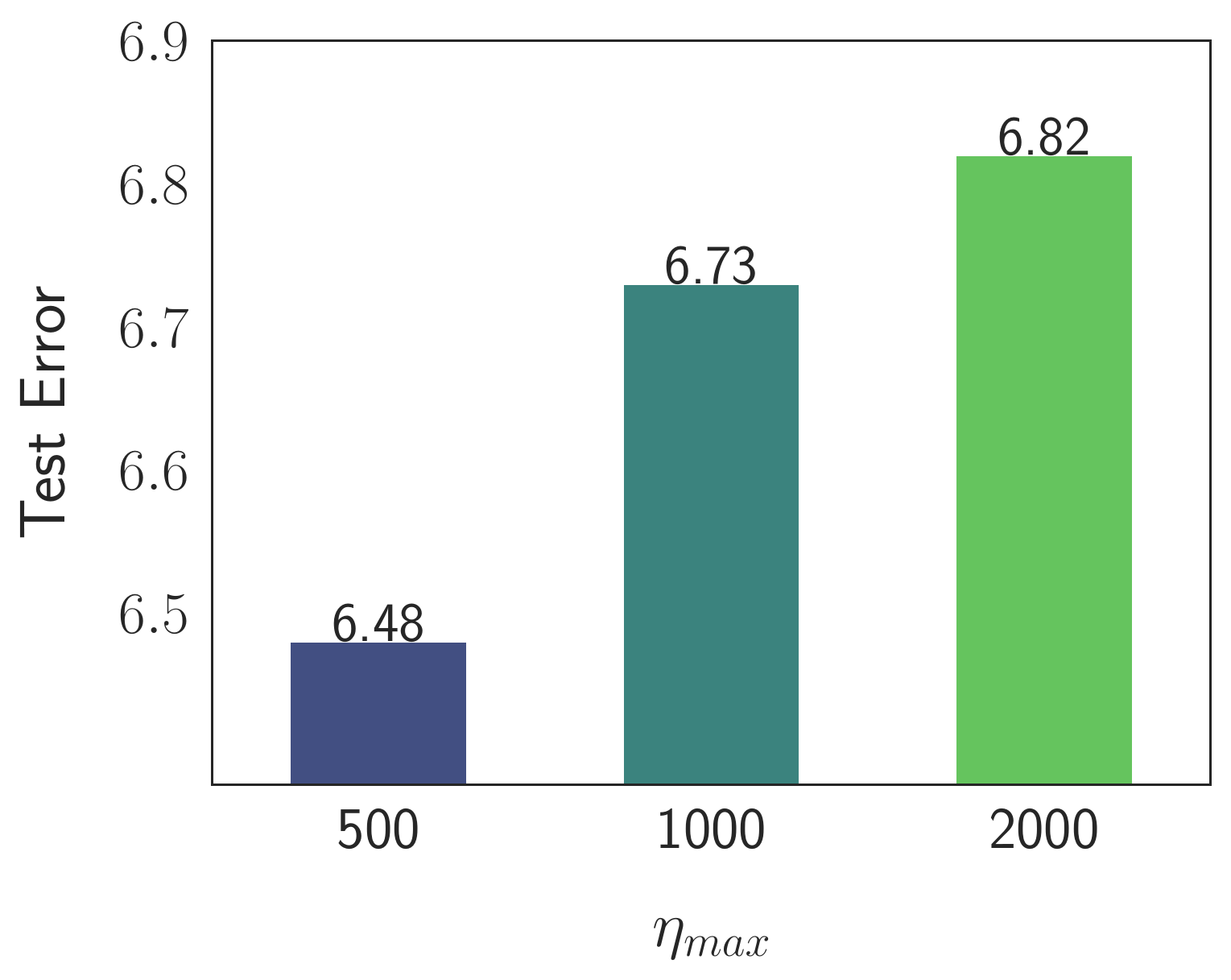}
\includegraphics[width=0.48\textwidth]{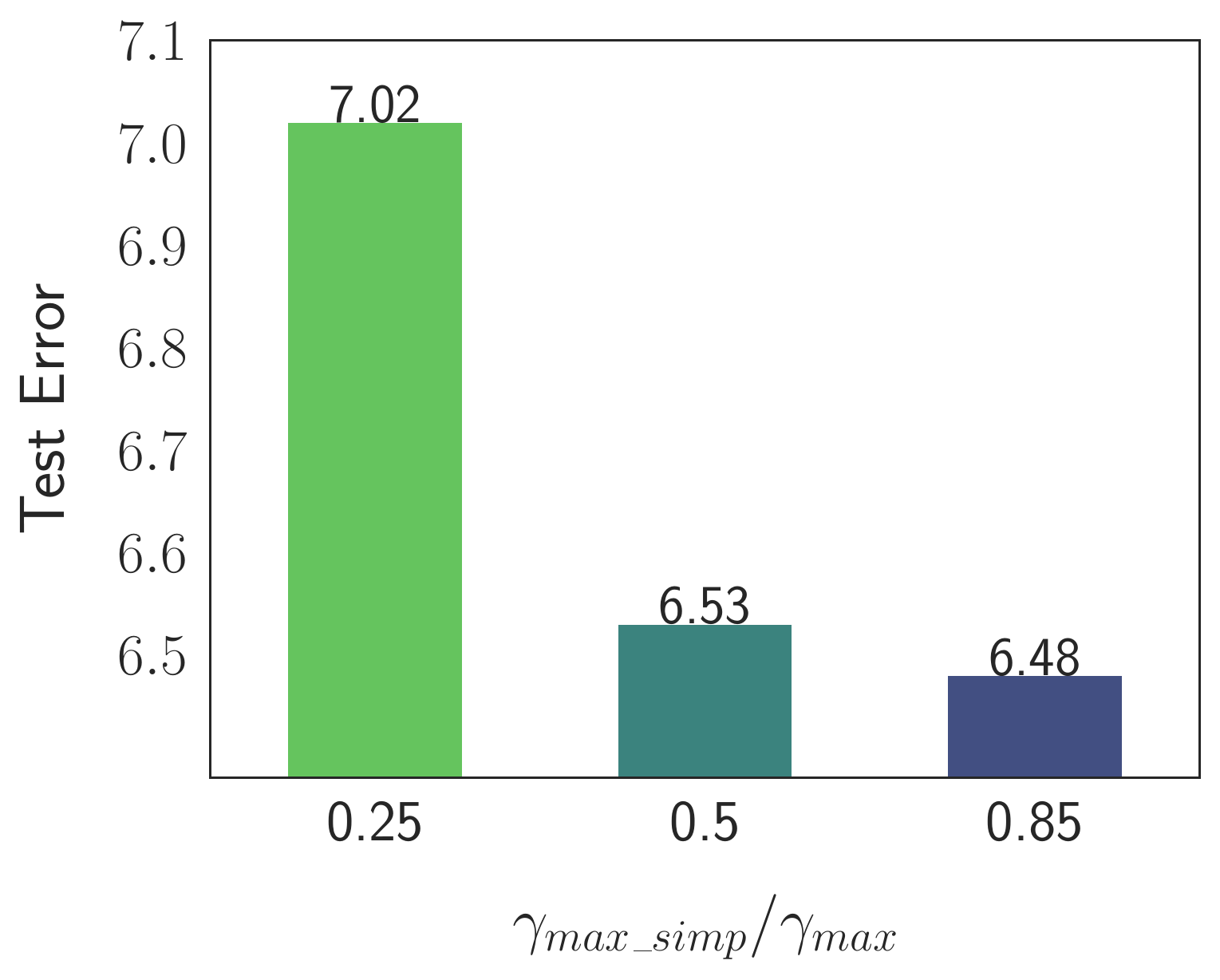}\\
\includegraphics[width=0.48\textwidth]{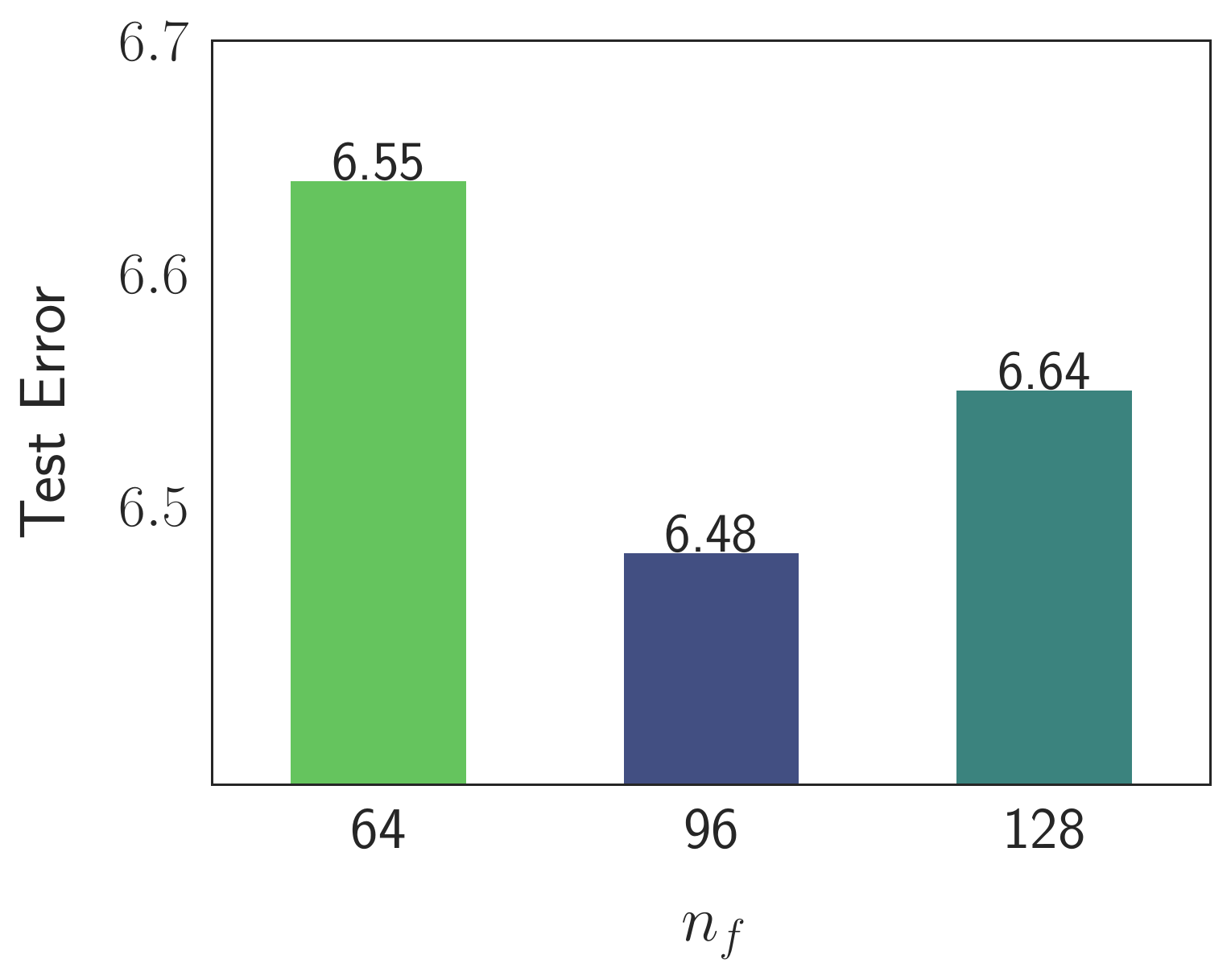}
\includegraphics[width=0.48\textwidth]{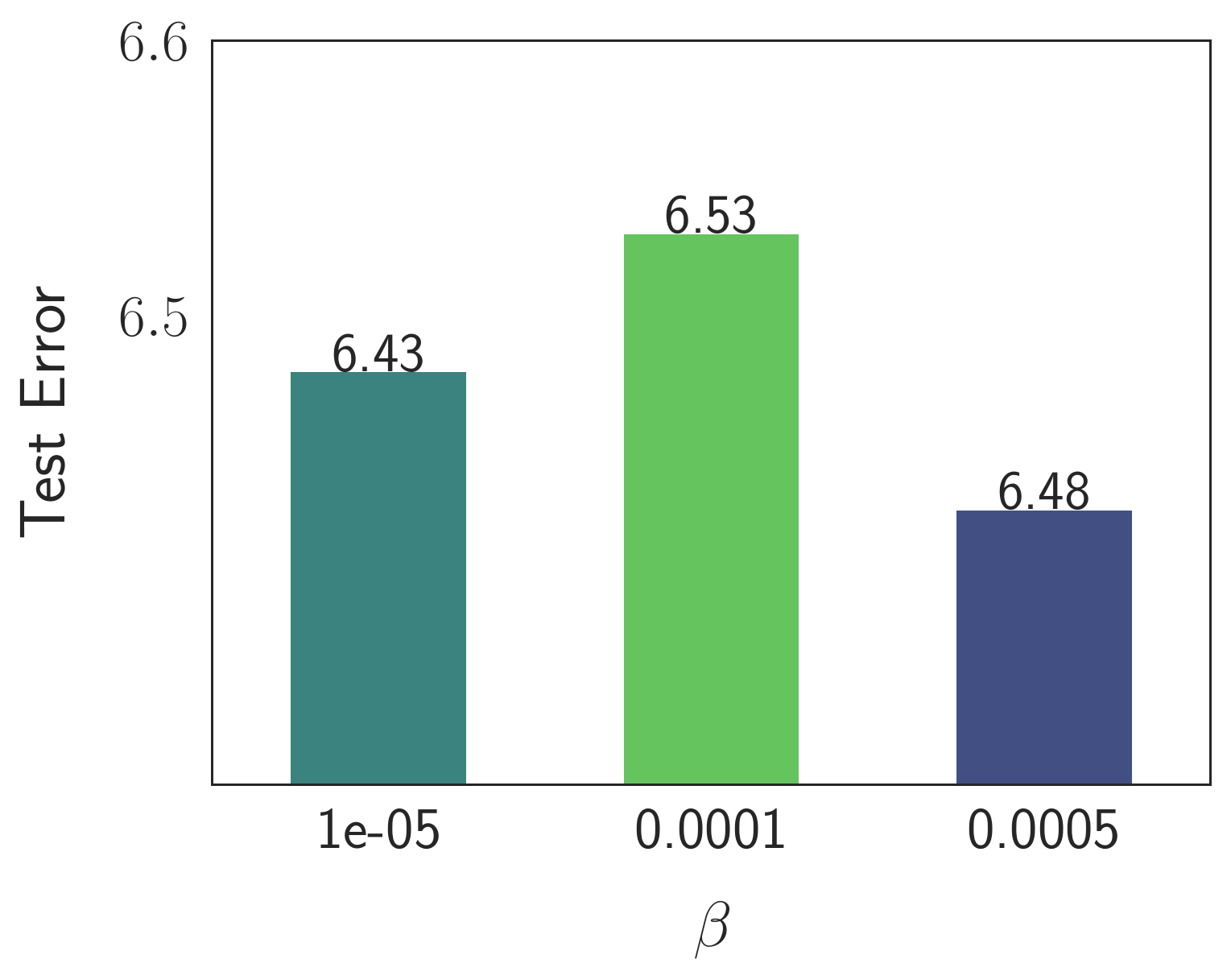}
\end{minipage}
\end{center}
\caption{Test error under different configuration of the NFT hyper-parameters, CNN-A architecture. }
\label{historight}
\end{figure}

We also evaluated the sensitivity of the system to some hyper-parameters of NFT, keeping the main network fixed. In Fig.~\ref{historight}, we report the test error of CNN-A, \textsc{mnist-back-image} dataset, for different configurations of $\eta_{max}$, $\frac{\gamma_{max\_{simp}}}{\gamma_{max}}$, $n_f$, $\beta$. In particular, after having selected a sample run that is pretty representative of the general trend we observed in the experiments, we changed one of the aforementioned parameters and computed the error rate.
Large values of $\eta_{max}$ reduce the freedom of auxiliary network in learning the transformation function.
Similarly, a short developmental plan with a small $\frac{\gamma_{max\_{simp}}}{\gamma_{max}}$ does not allow the main network to benefit from the progressively simplified data. 
In general, we did not experience a very significant sensitivity to the variations of $n_f$, and $64$ features turned out to be fine in most of the experiments, with some cases in which moving to $96$ was slightly preferable, as in the one we are showing in Fig.~\ref{historight}. Although in a fine-grained grid of values, we found that larger $\beta$ helped the auxiliary network to more quickly develop meaningful transformations. 
As a side note, we report that NFT was $\approx 1.5\times$ slower than CT, on average--see Sec.~\ref{sec:method}; performance optimization was outside the scope of this work.

\section{Conclusions and Future Work}
\label{sec:conclusions}
In this paper, we presented a novel approach to Friendly Training, according to which training data are altered by an auxiliary neural network in order to improve the learning procedure of a neural network-based classifier. Thanks to a progressive developmental plan, the classifier implicitly learns from examples that better match its current expectations, reducing the impact of difficult examples or noisy data during early training.
The auxiliary neural network is dropped at the end of the training routine. An extensive experimental evaluation showed that Neural Friendly Training leads to classifiers with improved generalization skills, overcoming vanilla Friendly Training in which an example-wise perturbation is estimated in an iterative manner.
Future work will focus on the investigation of different developmental plans and the evaluation of the impact of Neural Friendly Training in terms of robustness to adversarial examples.

\section*{Acknowledgements}
This work was partly supported by the PRIN 2017 project RexLearn, funded by the Italian Ministry of Education, University and Research (grant no. 2017TWNMH2).

\bibliography{biblio}

\end{document}